\documentclass[preprint,review,3p]{elsarticle}

\usepackage[colorlinks]{hyperref}
\usepackage[linesnumbered,boxed]{algorithm2e}
\usepackage{ulem,cleveref,amssymb, graphicx, threeparttable, booktabs, multirow, amsthm, amsmath,epic,arydshln,makecell,colortbl,subfigure}
\usepackage{breakurl}

\makeatletter
\AtBeginDocument{\Hy@breaklinkstrue}
\makeatother

\newdefinition{define}{Definition}[section]

\newcommand{\s}{\mathbf}
\newcommand{\w}{\widetilde}
\newcommand{\fs}{\footnotesize}

\definecolor{gray1}{gray}{0.56}%
\definecolor{gray11}{gray}{0.62}
\definecolor{gray2}{gray}{0.68}%
\definecolor{gray21}{gray}{0.74}
\definecolor{gray3}{gray}{0.80}%
\definecolor{gray31}{gray}{0.86}
\definecolor{gray4}{gray}{0.92}%

\begin{document}

\begin{frontmatter}



\title{Feature Selection with Redundancy-complementariness Dispersion}


 \author[author1,author2,author3]{Zhijun Chen}
 \ead{chenzj556@gmail.com}
 \author[author1,author2]{Chaozhong Wu}
 \ead{wucz@whut.edu.cn}
\author[author4,aa]{Yishi Zhang\corref{cor1}}
 \ead{zhang685@wisc.edu}
  \author[author5]{Zhen Huang}
 \ead{h-zhen@whut.edu.cn}
 \author[author3]{Bin Ran}
 \ead{bran@wisc.edu}
  \author[author1,author2]{Ming Zhong}
 \ead{mzhong@whut.edu.cn}
 \author[author1,author2]{Nengchao Lyu}
 \ead{lvnengchao@163.com}
\address[author1]{Intelligent Transport Systems Research Center, Wuhan University of Technology, Wuhan 430063, China}
\address[author2]{Engineering Research Center for Transportation Safety, Ministry of Education, Wuhan 430063, China}
\address[author4]{School of Management, Huazhong University of Science and Technology, Wuhan 430074, China}
\address[author5]{School of Automation, Wuhan University of Technology, Wuhan 430063, China}
\address[author3]{Department of Civil and Environment Engineering, University of Wisconsin-Madison, Madison, WI 53706, USA}
\address[aa]{Wisconsin School of Business, University of Wisconsin-Madison, Madison, WI 53706, USA}

\cortext[cor1]{Corresponding author at: Wisconsin School of Business, University of Wisconsin-Madison, 975 University Avenue, Madison, WI 53706, USA.}

\begin{abstract}
Feature selection has attracted significant attention in data mining and machine learning in the past decades.
Many existing feature selection methods eliminate redundancy by measuring pairwise inter-correlation of features, whereas the complementariness of features and higher inter-correlation among more than two features are ignored.
In this study, a modification item concerning the complementariness of features is introduced in the evaluation criterion of features. Additionally, in order to identify the interference effect of already-selected False Positives (FPs), the redundancy-complementariness dispersion is also taken into account to adjust the measurement of pairwise inter-correlation of features.
To illustrate the effectiveness of proposed method, classification experiments are applied with four frequently used classifiers on ten datasets. Classification results verify the superiority of proposed method compared with five representative feature selection methods.

\end{abstract}


\end{frontmatter}

\section{Introduction}\label{Introduction}
With the fast development of the world, the dimensional and size of data is fast-growing in most kinds of fields which challenge the data mining and machine learning techniques. Feature selection is an important and useful method that can effectively reduce the dimensionality of feature space while retaining a relatively high accuracy in representing the original data. Thus, it plays a fundamental role in many data mining and machine learning tasks, particularly in pattern recognition \cite{new12}, knowledge discovery \cite{new14,new15}, information retrieval \cite{new16,new17}, computer vision \cite{new18,new19}, bioinformatics \cite{new20}, and so forth. The effects of feature selection \cite{2} have been widely recognized for its abilities in facilitating data interpretation, reducing acquisition and storage requirements, increasing learning speeds, improving generalization performance, etc. Therefore, feature selection has attracted significant attention of more and more researchers \cite{JSandBEP,DEAFS,4,new24,new25}.

Generally speaking, the feature selection methods can be divided into two types: Wrapper and filter. Wrapper methods depend on specific learning algorithms. Thus the performance of wrapper methods is affected by the selected learning method. This may makes the wrapper methods computationally expensive in learning, since they must train and test the classifier for each feature subset candidate. Conversely, filter methods do not rely on any learning schemes. Instead, it is only based on some classifier-irrelevant metrics, including Fisher score \cite{6}, $\chi^2$-test \cite{8}, mutual information \cite{9,11,12,Zhangis_1,Zhangis_2}, Symmetrical Uncertainty(SU) \cite{13,DEAFS}, etc., to estimate the discrimination power of features. In this study, we only focus on filter methods.

Filter methods can also divided into feature subset selection and feature ranking ones, with regard to their searching strategy. The evaluation unit for subset selection methods is a set of features, thus the one with best discrimination power is trying to be discovered \cite{13,CFS,CMIM,IAMB2010,OnlineStreamFS}. Nevertheless, to find the best feature subset, $\mathrm{O}(2^m)$ candidate subsets (where $m$ is \# features in the original data) will be traversed for feature selection task cannot be solved optimally in polynomial-time unless $\s{P}=\s{NP}$ \cite{NPComplete}. Thus it is computationally intractable in practice. Unlike subset methods, feature ranking methods individually take features as the evaluation units and rank them according to their discrimination power \cite{Relief,Var_Of_relief,MIFS}. These methods usually employ heuristic search strategies such as forward search, backward search, and sequential floating search.

However, whatever feature ranking or feature subsets selection method, there are two problems possibly leading to the wrong ranking or lower capacity for classification. One is that the ignorance of feature interaction and dependencies may lead to redundancy, as some feature selection methods like MIM \cite{MIM} take the assumption of independence of features. For the real-world datasets, especially those high-dimensional ones, such strong assumption may produce results far from optimal. The other problem is that group capacity of features is usually ignored, since many methods only measure the relationship between any two features \cite{9,mRMR2,MIFS}. For example, a feature has low individual classification capacity and is highly dependent on other features may be overlooked and even misidentified as a redundant feature by only measuring its pairwise relationship with other features. However, since it is highly dependent on other features, it is also possible that the feature contributes largely to the discrimination power of the subset consisting of such features. Thus, it should be evaluated as a salient feature and selected. Since the dependence among features is related to both redundancy and complementariness, it is imperative to develop more precise correlation analysis in order to distinguish them effectively. To this end, we propose a novel feature selection algorithm which tries to modify the redundancy analysis applied in prior methods in this paper by introducing a modification item and a dynamic coefficient to effectively adjust the redundancy-complementariness identification process.

The remainder of the paper is organized as follows: Section \ref{Related work} reviews related work. Section \ref{Preliminaries} presents the Information theoretic metrics and evaluation criteria. A new feature selection method is included in section \ref{Framework}. In section \ref{Exp}, experimental study is described and the results are discussed. Finally, section \ref{conclusion} concludes this study and proposes possible further work.

\section{Related work}\label{Related work}
In recent decades, many kinds of feature selection methods have been studied. In general, there are two aims in these feature selection methods. One is to search the most class-relevant features, the other is to remove redundancy.
Most feature selection algorithms can effectively find relevant features \cite{new1,Var_Of_relief,new4}. A well-known example is Relief, which is developed by Kira and Rendell \cite{Relief}. The main idea of Relief is to rank features in terms of the weight corresponding to their ability to both discriminate instances with different class labels and cluster those with same class labels based on the distance between instances. However, Relief method may be ineffective since similar weights of two or more features cannot be removed by this method. In other words, this implies that redundant features cannot be identified. A typical and widely used extension of Relief is ReliefF \cite{ReliefF}, which is competent to the noisy and incomplete datasets. However, it is still unable to remove redundant features.
Redundant features are considered to have negative effects on the accuracy and speed of classification methods, hence many feature selection methods are proposed to address this problem by statistic-based merics \cite{Kohavi_John,KollerSahami,MIFS,CFS,13}. For example, Correlation based Feature Selection (CFS) algorithm proposed by Hall et al. \cite{CFS} adopts $cor$ value to simultaneously measure a feature subset's correlation to the class and inter-correlation among features in it. CFS selects the subset which obtains the maximum $cor$ value. However CFS does not designate specific search approaches, thus how to select feature subsets still remains to be a problem.

Minimum Redundancy and Maximum Relevance (mRMR) criterion and its variants \cite{9,mRMR2,mRMR14,MIFS} apply information theoretic metrics to separately measure class-relevance and pairwise correlation between features. A comprehensive score consisting of the two indices is applied to evaluate and select features. Fast Correlation Based Feature selection algorithm (FCBF) proposed by Yu and Liu \cite{13} is another typical method that separately handles relevance and redundancy. FCBF utilizes Symmetrical Uncertainty ($SU$) as the merci to represent class-relevance and pairwise correlation. If the class-relevance of a feature is lower than that of another and the correlation between them, it would be identified as a redundant features and thus to be removed. Recently, an extenuation of FCBF was proposed in order to identify redundant features more precisely \cite{Fast-FCBF}. All of the above mentioned methods take pairwise correlation as the redundancy index and identify features with high such index to be redundant, while ignoring 1) complementary correlation between features (which we will discuss detailed in section \ref{DRM}) and 2) correlation among more than two features, which still remain to be problems that impair the performance of feature selection.

 Much effort has been made to tackle the former problem mentioned above \cite{CMIM,CMIM_WANG,IAMBs,CCR,JMI,Zhangis_1,Zhangis_2,DEAFS,DISR,MIFS-U}. Flueret \cite{CMIM} and Wang et al. \cite{CMIM_WANG} propose Conditional Mutual Information Maximization (CMIM) criterion for feature selection. CMIM harnesses Conditional Mutual Information (CMI) to measure the intensity of relevance and redundancy since CMI can implicitly identify complementary correlation between features, i.e. a large value of $CMI(F ; C | \w{F})$ implies 1) $F$ is relevant to class $C$, and 2) $F$ is highly complementary with $\w{F}$, many information theoretic feature selection methods apply it to build up their evaluation criteria \cite{mRR,CCR,IAMBs,Fast_IAMB}. There are also several methods explicitly identifying redundancy and complementary correlation without CMI. Algorithms based on Joint Mutual Information (JMI) criterion \cite{JMI,DISR} take into account mutual information between pairs of features and class. Since the feature relevant to class and the one complementary to salient features will obtain high JMI values, they both will be identified as salient features and thus is more possible to be selected. Although the above mentioned methods try to recognize complementariness from the pairwise correlation of features, measuring pairwise correlation is actually an approximation to measuring the correlation among more than two features. Under this circumstance, features that is strongly complementary to the certain selected feature(s) but not significantly correlated with the feature group are possible to be selected using such approximation, which will in turn intervene the later selection process.

\section{Information theoretic metrics and evaluation criteria}\label{Preliminaries}

\subsection{Entropy, mutual information, and conditional mutual information}\label{MI}
In this section, some essential information theoretic metrics used in our method will be described. The entropy, a fundamental unit of information, is used to quantify the uncertainty preset in the distribution of $X$, which is formed as
 \begin{equation*}
  H(X)=
  -\sum_{x\in X}p(x)\log p(x),
\end{equation*}
where $x\in X$ denotes the possible value assignments of $X$, $p(x)$ is the distribution of $x$ (for convenience, we hereafter use the notation $\log$ to denote the base 2 logarithm instead of $\log_2$). According to the probability theory, one can use conditional entropy to  quantify the uncertainty one variable conditioned on another one. The conditional entropy of $X$ given $Y$ is defined as
\begin{equation*}
  H(X|Y)=
  -\sum_{y\in Y}\sum_{x\in X}p(xy)\log p(x|y),
\end{equation*}
Mutual Information (MI) between two random variables $X$ and $Y$ can be described as follows
\begin{equation*}
  I(X;Y)=
  \sum_{x\in X}\sum_{y\in Y}p(xy)\log\frac{p(xy)}{p(x)p(y)},
\end{equation*}
 where $x\in X$ and $y\in Y$ are the possible value assignments of $X$ and $Y$, respectively. MI can be considered as the amount of information shared by two variables. In feature selection field, it is one of the most widely used metrics for measuring the correlation intensity of two features. Note that the MI is a symmetric merci, i.e. $I(X;Y)=I(Y;X)$. $I(X;Y) = 0$ implies that $X$ and $Y$ are statistically independent. Conditional mutual information (CMI), which is an extension of MI for measuring the conditional dependence between two random variables given the third, is defined as
\begin{equation*}
  I(X;Y|Z)= \sum_{z\in Z}p(z)\sum_{x\in X}\sum_{y\in Y}p(xy|z)\log\frac{p(xy|z)}{p(x|z)p(y|z)}.
\end{equation*}

$I(X;Y|Z)$ can be interpreted as the information shared between $X$ and $Y$ given the value of a third variable ($Z$). MI and CMI can also be expressed with entropies as follows:
\begin{equation*}
    I(X;Y)=H(X)-H(X|Y)
\end{equation*}
and
\begin{equation*}
    I(X;Y|Z)=H(X|Z)-H(X|Y,Z).
\end{equation*}

\subsection{Relevance, redundancy, and complementariness analysis}\label{DRM}

The motivation of using MI to solve feature selection problem is that a larger MI between the feature and class should imply a potentially greater discrimination ability when using the feature. In addition, a commonly cited justification for using MI in feature selection is that MI can be used to write both an upper and lower bound on the Bayes error rate \cite{xin33}. It can simply be applied as the criterion of a filter taking the form of
\begin{equation}\label{MIM}
	J(F)=I(F;C),
\end{equation}
where $J(\cdot)$ denotes the evaluation criterion, $F$ denotes a candidate feature and $C$ denotes the class. Intuitively, the top $m$ candidates which maximize $J(\cdot)$ could be selected, where $m$ is a predefined number or decided by some stop criterion. In fact, this criterion takes the assumption that each feature is independent to all other features, which makes the criterion very efficient.
However, such an assumption is so strong in practice that almost all the features may be mutually dependent to others, which makes the criterion shown in Eq.\eqref{MIM} be far from optimal. In general, it is widely recognized that a salient set of features should not only be individually relevant to class, but also should not be redundant to other features in the set. In order to identify redundancy, mRMR and its variants are proposed which can be generally formed as
\begin{equation}\label{mRMR}
J(F)=D(F)-R(F)
\end{equation}
where $D(F)$ represents relevance between $F$ and class $C$, $R(F)$ describes redundancy between $F$ and the selected features in the subset $\s{S}$. Usually, like in mRMR \cite{9}, $D(F)$ and $R(F)$ take the forms of MI. This criterion can efficiently find the features with high class-relevance and low dependence with respect to each other in $\s{S}$. However, term ``redundancy'' not only implies that features are highly dependent to each other, but also indicates which one would be substitutable, i.e. their discrimination power would be significantly impaired when some other feature(s) are(is) given. From this viewpoint, only considering dependence between features is not enough to effectively identify redundancy. In other words, a feature which is dependent on another may not definitely imply to be redundant. Instead, the two features may complementary to each other, i.e they would have stronger discriminatory power as a group (but may weak as individuals), particularly in microarray data analysis \cite{xin31,new6}. To this end, a complementary modification item is introduced as
\begin{equation}\label{fou}
J(F)=D(F)-(R(F)-M(F))
\end{equation}
where $M(F)$ is an item to identify complementary correlation between $F$ and selected features in $\s{S}$. In the context of MI, if $R(C)$ takes the form of $\sum_{F_s\in\s{S}}I(F;F_s)$ (as in mRMR), $M(F)$ could thus be denoted as $\sum_{F_s\in\s{S}}I(F;F_s|C)$, which represents the information shared between $F$ and $F_s$ given class $C$. In order to illustrate this, we first show the relationship between $R(F)$ and $M(F)$ as follows
\begin{eqnarray}\label{R-M=MI-CMI}
R(F) - M(F) & = & I(F;F_s)- I(F;F_s|C) \nonumber\\
& = &\displaystyle\sum_{f_s\in F_s}\sum_{f\in F}p(ff_s)\log\frac{p(ff_s)}{p(f)p(f_s)} - \sum_{c\in C}p(c)\sum_{f_s\in F_s}\sum_{f\in F}p(ff_s|c)\log\frac{p(ff_s|c)}{p(f|c)p(f_s|c)} \nonumber\\
& = & \sum_{c\in C}\sum_{f\in F}\sum_{f_s\in F_s}p(ff_sc)\log\left(\frac{p(ff_s)}{p(f)p(f_s)}\cdot\frac{p(f|c)p(f_s|c)}{p(ff_s|c)}\right)  \nonumber \\
& = & \sum_{c\in C}\sum_{f\in F}\sum_{f_s\in F_s}p(ff_sc)\log\frac{p(ff_s)p(fc)p(f_sc)}{p(f)p(f_s)p(c)p(ff_sc)}  \nonumber \\
& = & \sum_{c\in C}\sum_{f\in F}\sum_{f_s\in F_s}p(ff_sc)\log\left(\frac{p(fc)}{p(f)p(c)}\cdot\frac{p(ff_s)p(f_sc)}{p(ff_sc)p(f_s)}\right)  \nonumber \\
& = &\displaystyle\sum_{f\in F}\sum_{c\in C}p(fc)\log\frac{p(fc)}{p(f)p(c)} - \sum_{f_s\in F_s}\sum_{f\in F}\sum_{c\in C}p(ff_sc)\log\frac{p(fc|f_s)}{p(f|f_s)p(c|f_s)} \nonumber\\
& = & I(F;C) - I(F;C|F_s).
\end{eqnarray}
We now explain $R(F)-M(F)$ using Eq.\eqref{R-M=MI-CMI}, since the relationship between $I(F;C)$ and $I(F;C|F_s)$ is straightforward: If $I(F;C)$ is much great than $I(F;C|F_s)$, the relevance between $F$ and class $C$ would become significantly weak after given the information of $F_s$. In other words, $F$ is redundant to $F_s$. Conversely, if $I(F;C)$ is much small than $I(F;C|F_s)$, the relevance between $F$ and class $C$ would become significantly strong after given the information of $F_s$. i.e. $F$ is complementary to $F_s$. Thus, $R(F)-M(F)$ could be applied to simultaneously measure redundancy and complementary correlation: When $R(F)-M(F)>0$, it captures the magnitude of redundancy between $F$ and $F_s$; when $R(F)-M(F) < 0$, it captures the magnitude of complementary correlation between $F$ and $F_s$. In the context of MI, the following expression could be applied to be the evaluation criterion according to Eq.\eqref{fou}
\begin{equation}\label{proposed_main}
	J(\cdot)=I(F;C)-Pair\_Cor(F;\s{S})
\end{equation}
where $Pair\_Cor(F;\s{S})$ takes the form of
\begin{equation}\label{pair_cor_temp}
	Pair\_Cor(F;\s{S})=\sum_{F_s\in\s{S}}\left(I(F;F_s)-I(F;F_s|C)\right).
\end{equation}
For the sake of convenience for the discussion in the following sections, we denote $cor(F;F_s) = I(F;F_s)-I(F;F_s|C)$ and thus Eq.\ref{pair_cor_temp} can be rewriten as
\begin{equation}\label{pair_cor}
Pair\_Cor(F;\s{S})=\sum_{F_s\in\s{S}}cor(F;F_s).
\end{equation}
It is noted that although Eq.\eqref{pair_cor} can measure both redundancy and complementary correlation, it is still a pariwise-based criterion since it only catches the relationship between two features. Criteria that only concern pairwise correlation among features is also called first-order approximation in literature \cite{FEAST}. We will further discuss the limitation of Eq.\eqref{pair_cor} in detail in the next section.

\section{Feature selection with redundancy-complementariness dispersion}\label{Framework}
\subsection{Interference effect of false positives}

First-order approximation is a prevailing strategy that seems to bring the best trade-off between executional efficiency and the selected features quality. Yet ignoring the group effect of features is still known to be suboptimal although taking the pairwise relevance effect into account. As mentioned before, feature selection methods that only handle individual relevance take the assumption of mutual independence among features. Similarly, first-order approximation in redundancy analysis only concentrates on individual redundancy. In other words, it takes the assumption that all the selected features are mutually independent. Since the first-order approximation only identify pairwise correlation, it is not able to take high inter-feature correlation into account, thus may misidentify and select actually-redundant features (i.e. False Positives, which is denoted as FPs hereafter in the paper; Similarly, we use the term True Positives (TPs) to denote the selected actually-salient features hereafter in the paper), which will in turn intervene the later selection process.

\begin{figure*}[!ht]
	\begin{center}
		\includegraphics[trim = 13em 47em 20em 20em, scale = 1.1]{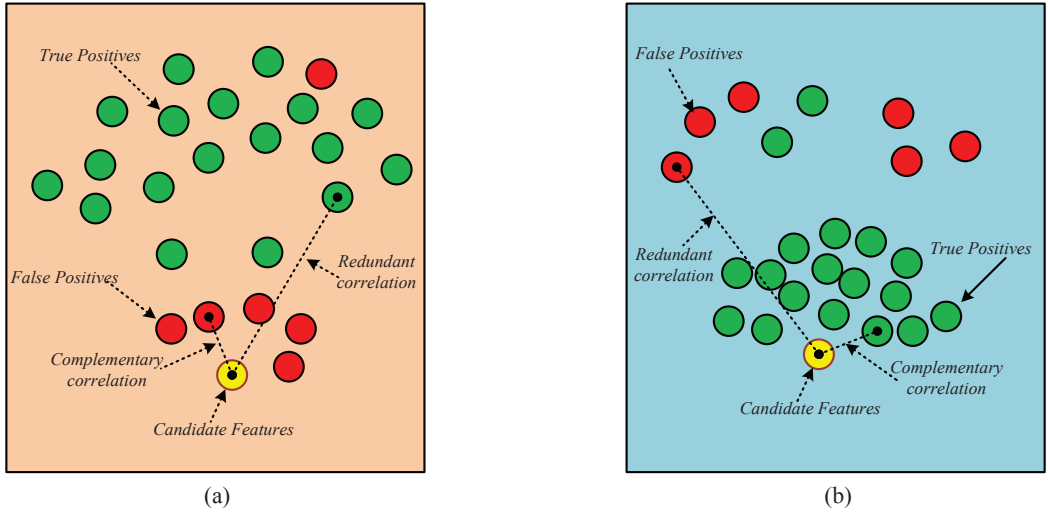}\\
		\caption{Toy examples of interference effect of FPs.}\label{density_example}
	\end{center}
\end{figure*}

More specifically, only focusing on pairwise correlation may give chance to FPs to intervene the evaluation of candidates. Suppose the selected feature subset already contains FPs, the pairwise correlation between the candidate and each FP is an interference that prompts the candidate to be given unduly high status if such correlation is influential to the value of the evaluation criterion $J(\cdot)$. recall that the correlation between candidate and each selected features is denoted as $cor(F;F_s)$ where $F_s\in\s{S}$ ($\s{S}$ is the selected feature subset) and thus $Pair\_Cor(F;\s{S})=\sum_{F_s\in\s{S}}cor(F;F_s)$, the interference effect of FPs can be illustrated in two possible scenarios shown in Figs.\ref{density_example} (a) and (b), where node in yellow, nodes in red, and nodes in green denote the candidate, FPs, and TPs, respectively. Distance between yellow node and any other node is in proportion to the strength of their pairwise correlation, e.g. a short distance corresponds to the complementary correlation, while long corresponds to the redundant correlation.

{\it Scenario} 1: FPs are close to the candidate. As shown in Fig.\ref{density_example} (a), most of TPs are distant to the candidate, which implies that the candidate is more likely to be redundant rather than complementary to FPs (which corresponds to positive $cor$ value in terms of Eq.\eqref{R-M=MI-CMI}) and thus it is possibly a redundant feature. However, as FPs are very close to the candidate, they are more likely to be complementary and the corresponding $cor$ value tend to be negative. Under this circumstance, the complementary correlation between candidate and FPs impairs the reliability of the estimation of $Pair\_Cor(F;\s{S})$ and thus makes the candidate to be overestimated.

{\it Scenario} 2: FPs are distant to the candidate. Fig.\ref{density_example} (b) shows that most of TPs are close to the candidate. This implies that the candidate is more complementary to TPs and thus more likely to be a salient feature that should be selected. However, it is redundant to the distant FPs and the corresponding $cor$ value tend to be positive, thus also impairs the reliability of the estimation of $Pair\_Cor(F;\s{S})$ and makes the candidate to be underestimated.

Actually, the interference effect of FPs revealed in the above scenarios can be depicted by the dissimilarity of the selected features. That is, the intensity of the interference effect of FPs depends on the amount of the dispersion of the correlation between candidate and the selected features. When a certain value of $Pair\_Cor(F;\s{S})$ is given, the correlation between $F$ and FPs in $\s{S}$ which is more likely to be complementary corresponding to larger negative $cor$ values would lead to the correlation between $F$ and TPs in $\s{S}$ which is more likely to be redundant corresponding to larger positive $cor$ values, and vice versa. We call such dissimilarity as redundancy-complementariness dispersion. As a heuristic, we apply standard deviation of $cor$ to capture such dispersion in order to possibly identify the interference effect of FPs, for standard deviation is always the best index for risk estimation and instability identification. The standard deviation of $cor(F;F_{s})$ given the selected feature subset $\s{S}$ takes the form of
\begin{equation}\label{std}
\sigma(F;{\s{S}})=\left(\frac{\sum_{F_s\in\s{S}}\left(cor(F;F_s)-\mu(F;\s{S})\right)^2}{|\s{S}|}\right)^{\frac{1}{2}},
\end{equation}
where $\mu(F;\s{S})$ is the mean value of $cor(F;F_s)$ calculated as
\begin{equation}
\mu(F;\s{S})=\frac{Pair\_Cor(F;\s{S})}{|\s{S}|}.
\end{equation}
Thus, the smaller the value of $\sigma(F;{\s{S}})$, the less influential the interference effect of FPs. We try to find salient candidates not only with more complementariness and less redundancy, but also less redundancy-complementariness dispersion, i.e. a small value of $\sigma(F;\s{S})$, to heuristically avoid the interference effect of FPs. To this end, we use $\sigma(F;\s{S})$ to adjust the value of $Pair\_Cor$. Recall that $Pair\_Cor$ simultaneously measures two types of correlation, i.e. redundancy (where the value of $Pair\_Cor$ is positive) and complementariness (where the value is negative). Taking this into account, we use the following criterion
\begin{equation}\label{J}
J_{RID}=D(F;C)-\phi(F;\s{S})\cdot Pair\_Cor(F;\s{S})
\end{equation}
where
\begin{equation}\label{coefficient}
\phi(F;\s{S}) = \begin{cases}
1+\sigma(F;\s{S}) &\quad\quad Pair\_Cor(F;\s{S}) \ge 0 \\
1-\sigma(F;\s{S}) &\quad\quad Pair\_Cor(F;\s{S}) < 0
\end{cases}
\end{equation}
to evaluate and select features among candidates. Note that $\phi(F;\s{S})$ is defined piecewise for different types of correlation. Also, we use $1+\sigma(F;\s{S})$ and $1-\sigma(F;\s{S})$ rather than $\sigma(F;\s{S})$ as the coefficient of $Pair\_Cor(F;\s{S})$ in order to reduce the estimation bias for $\sigma(F;\s{S})$ particularly when there are only a few features selected in $\s{S}$.

\subsection{Proposed method}
\begin{figure*}[!ht]
	\begin{center}
		\includegraphics[trim = 9em 43em 20em 31em, scale = 1.2]{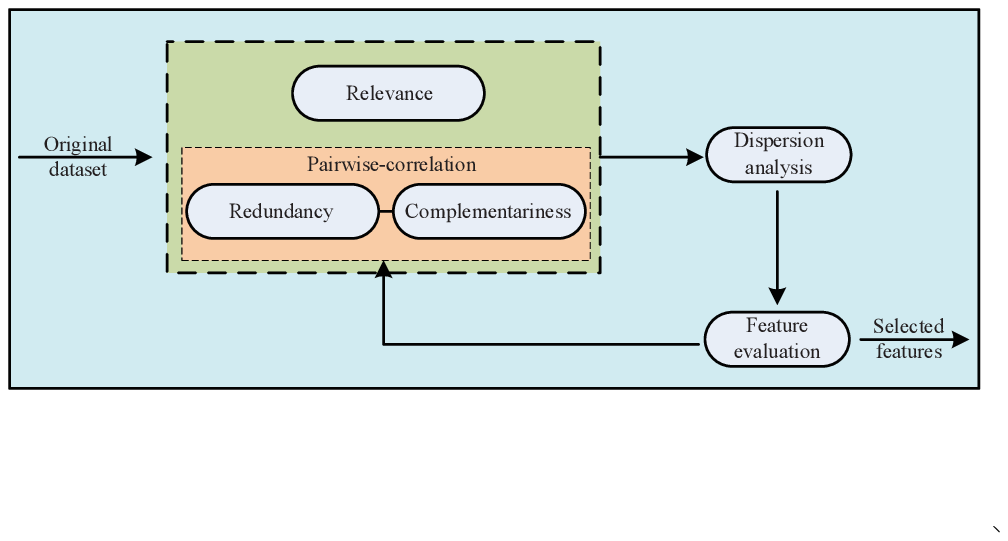}\\
		\caption{A new framework for feature selection.}\label{framework}
	\end{center}
\end{figure*}

Based on the above analysis, we propose our feature selection framework shown in Fig.\ref{framework}. It not only consider class-relevance and pairwise inter-correlation of features, but also take into account the effect of redundancy-complementariness dispersion. Similar to most of the feature selection methods, proposed method also applies a sequential forward searching strategy to select features. That is, only one candidate would be selected at each iteration.

We show the pseudo code of proposed algorithm in Algorithm \ref{alg}.
\begin{algorithm}[!th]
	\caption{RCDFS: {\bf R}edundancy-{\bf C}omplementariness {\bf D}ispersion-based {\bf F}eature {\bf S}election}
	\label{alg}
	\KwIn{ $\s{D}$ /*dataset*/, $\s{F}$ /*feature set*/, $C$ /*class*/, $\delta$ /*expected \# features to be selected*/}
	\KwOut{$\s{S}$ /*selected feature subset*/}
	Initialize $\s{S}\leftarrow\varnothing$, $k\leftarrow 1$  \\
	\Repeat{$k \ge \delta$}{
		\ForEach{$F\in\s{F}$}{
			$Relevance \leftarrow I(F;C)$  \\
			$Pair\_Cor\leftarrow 0$ \\
			\ForEach{$F_s\in\s{S}$}{
				$cor \leftarrow I(F;F_s) - I(F;F_s|C)$  \\
				$Pair\_Cor \leftarrow Pair\_Cor + cor$   \\
			}
			Calculate $\sigma(F;\s{S})$ according to Eq.\eqref{std}\\
			\eIf{$Pair\_Cor \ge 0$}{$\phi \leftarrow 1+\sigma(F;\s{S})$}{$\phi \leftarrow 1-\sigma(F;\s{S})$}
			$J(F) \leftarrow Relevance - \phi\cdot Pair\_Cor $ \\
		}
		$\s{S}\leftarrow\s{S}\cup\{\w{F}\}$ satisfying $\w{F} = \arg\max_{F\in\s{F}} J(F)$ \\
		$\s{F}\leftarrow\s{F}-\{\w{F}\}$ \\
		$k\leftarrow k + 1$\\
	}
	\Return{$\s{S}$}
\end{algorithm}

Algorithm \ref{alg} contains a `repeat' loop and two `for' loops and a calculation process of $\sigma(F;\s{S})$ (line 11 in Algorithm \ref{alg}) which takes at least $|\s{S}|$ loops for the calculation. Thus the iteration complexity of Algorithm \ref{alg} is $\mathrm{O}(\delta\cdot|\s{F}|^2)$, where $\delta$ is the predefined number of selected features. Since there is only one candidate to be selected at the end of each iteration when traversing $F_s\in\s{S}$, we only need to get the additional information of the newly-added feature rather than traversing $\s{S}$ again. As for the calculation of $\sigma(F;\s{S})$ (i.e. the variance of $cor$), we could taking an alternative formulation of variance, i.e. $Var(X)=E(X^2)-E^2(X)$ to make use of the loops in Algorithm \ref{alg}. That is, to get $\sigma(F;\s{S})$ , we designate $P$ to record the summation of $cor^2$ and $Q$ to record the summation of $cor$, and then we have
$$
\sigma(F;\s{S}) = \left(\frac{P - Q^2/|\s{S}|}{|\s{S}|}\right)^{\frac{1}{2}}.
$$
Taking the above into account, we show the fast implementation of Algorithm \ref{alg} as Algorithm \ref{fast}.

\begin{algorithm}[!th]
	\caption{A fast implementation of RCDFS}
	\label{fast}
	\KwIn{ $\s{D}$ /*dataset*/, $\s{F}$ /*feature set*/, $C$ /*class*/, $\delta$ /*expected \# features to be selected*/}
	\KwOut{$\s{S}$ /*selected feature subset*/}
	Initialize $\s{S}\leftarrow\varnothing$, $F_{new}\leftarrow\varnothing$, $\Delta(F)\leftarrow 0$ for $\forall F\in\s{F}$, $Pair\_Cor(F)\leftarrow 0$ for $\forall F\in\s{F}$, $k\leftarrow 0$  \\
	\ForEach{$F\in\s{F}$}{
		$Relevance(F) \leftarrow I(F;C)$  \\
	}
	$F_{new}\leftarrow\w{F}$ satisfying $\w{F} = \arg\max_{F\in\s{F}} Relevance(F)$ \\
	$\s{S}\leftarrow\s{S}\cup\{F_{new}\}$\\
	$\s{F}\leftarrow\s{F}-\{F_{new}\}$ \\
	$k\leftarrow k + 1$ \\
	\Repeat{$k \ge \delta$}{
		\ForEach{$F\in\s{F}$}{
			$Relevance \leftarrow I(F;C)$  \\
			$cor \leftarrow I(F;F_{new}) - I(F;F_{new}|C)$  \\
			$\Delta(F) \leftarrow \Delta(F) + cor^2$ \\
			$Pair\_Cor(F) \leftarrow Pair\_Cor(F) + cor$   \\		
			$\sigma(F;\s{S})\leftarrow\left(\frac{\Delta(F) - Pair\_Cor(F)^2/|\s{S}|}{|\s{S}|}\right)^{\frac{1}{2}}$  \\
			\eIf{$Pair\_Cor(F) \ge 0$}{$\phi \leftarrow 1+\sigma(F;\s{S})$}{$\phi \leftarrow 1-\sigma(F;\s{S})$}
			$J(F) \leftarrow Relevance(F) - \phi\cdot Pair\_Cor(F) $ \\
		}
		$F_{new}\leftarrow\w{F}$ satisfying $\w{F} = \arg\max_{F\in\s{F}} J(F)$ \\
		$\s{S}\leftarrow\s{S}\cup\{F_{new}\}$\\
		$\s{F}\leftarrow\s{F}-\{F_{new}\}$ \\
		$k\leftarrow k + 1$ \\
	}
	\Return{$\s{S}$}
\end{algorithm}
By utilizing the additional information gained at the latest iteration, the complexity of Algorithm \ref{fast} reduces to  $\mathrm{O}(\delta\cdot|\s{F}|)$, which is more efficient than Algorithm \ref{alg}. Thus, we implement proposed method according to Algorithm \ref{fast} in the experiments to verify the performance of RCDFS.

\section{Experiment study}\label{Exp}

In order to evaluate the performance and effectiveness of proposed method, the most representative and well-performed feature selection methods (CMIM \cite{CMIM}, mRMR \cite{9}, FCBF \cite{13},MIM \cite{MIM} and ReliefF \cite{ReliefF}) are used to compare with proposed algorithm. The brief reviews on above five selected feature selection algorithm are described as follows.
\begin{itemize}
 \item CMIM (Conditional Mutual Information Maximization) \cite{CMIM}: This well-known algorithm makes use of CMI to simultaneously measure class-relevance and inter-correlation of features, applying the following function
  $$J(F)=\displaystyle\min_{\w{F}\in\s{S}} I(F;C|\w{F})$$
  as the evaluation criterion, taking the heuristic that $\w{F}$ satisfying $\min_{\w{F}\in\s{S}} I(F;C|\w{F})$ could best represent the conditioning set $\s{S}$.
 \item mRMR (minimum Redundancy and Maximum Relevance) \cite{9}: It is a very famous feature selection algorithm that uses MI to measure class-relevance and pairwise dependence. It selects feature satisfying
 $$\displaystyle J(F)=I(F;C)-\frac{1}{|\s{S}|}\sum_{F_s\in\s{S}}I(F;F_s)$$
 in a greedy manner, where $I(F;C)$ measures the class-relevance of $F$ and $\frac{1}{|\s{S}|}\sum_{F_s\in\s{S}}I(F;F_s)$ measures the average pairwise dependence between $F$ and $F_s\in\s{S}$. Note that we have already introduced it in section \ref{DRM}.
 \item FCBF (Fast Correlation-Based Feature selection) \cite{13}: In this algorithm, Symmetrical Uncertainty (SU) is used as the evaluation merci. It first ranks features in descending order. Then it eliminates redundant features in terms of an approximate Markov blanket criterion: If $SU(F_1;C)>SU(F_2;C)$ and $SU(F_1;C)>SU(F_1;F_2)$, $F_2$ is thus identified as a redundant feature of $F_1$ and thus would be eliminated. For this method, we set the predefined threshold $\gamma = 0$ as suggested by \cite{13}.
 \item  MIM (Mutual Information Maximization) \cite{MIM}: It is the most basic feature ranking algorithms based on mutual information that only concerns the class-relevance of features. We have also introduced it in section \ref{DRM}. It applies$$J(F) = I(F;C)$$as the criterion to select the top $m$ features with the highest value of $I(F;C)$. It is one of the most typical benchmark algorithms in the field of feature selection.
 \item ReliefF \cite{ReliefF}: It is a well-known distance-based feature ranking method that searches nearest neighbors of samples for each class label and then weights features in terms of how well they differentiate samples for different class labels. As for the parameter settings, we use 5 neighbors and 30 instances throughout the experiments as suggested by \cite{ReliefF}.
\end{itemize}

Weka (Waikato environment for knowledge analysis) \cite{Weka} is chosen as the classification platform. Since FCBF, MIM, and ReliefF have already been integrated in Weka, we directly use them to generate datasets with their selected features before classification. CMIM, mRMR, and the proposed method are implemented in Java and with Weka interfaces. All experiments are conducted on a 2.60 GHz CPU, 8GB RAM personal computer with Windows 7.

\subsection{Datasets}
In order to validate the performance of the proposed method, ten frequently used datasets are applied in our experiments, where six of them (mushroom, kr-vs-kp, sonar, multiple features ka, DNA, and isolet5) are well known UCI datasets and the rest (Colon Tumor, BCR\_ABL, Prostate Cancer, and Breast Cancer) are gene microarray datasets with high dimensionality (i.e. containing more than 2000 features). General information of these datasets are summarized in Tab.\ref{datasets}. For the continuous and mixed datasets, a supervised discretization method called MDL \cite{MDL} is employed to discrete continuous features before feature selection and classification.
\begin{table}[!ht]
\centering
\footnotesize
    \begin{threeparttable}
    \caption{Description of datasets}\label{datasets}
        \begin{tabular}{rl r@{}lr@{}lcl}
            \toprule
                 \#\phantom{.000}  & Name  & \multicolumn{2}{l}{\phantom{}\# samples\phantom{00}} & \multicolumn{2}{l}{\phantom{}\# features\phantom{0}} & Type & \# classes \\
            \midrule
           1\phantom{0000}&     mushroom    &  \phantom{00} 8124    &    &  \phantom{00}22      &  & nominal  &  \phantom{0000}2\\
           2\phantom{0000}&     kr-vs-kp    &  \phantom{00} 3196    &    &  \phantom{00} 36     &  & nominal  &  \phantom{0000}2\\
           3\phantom{0000}&     sonar   &  \phantom{00} 208     &    &  \phantom{00} 60         &  & nominal  &  \phantom{0000}2\\
           4\phantom{0000}& multiple features kahunen   &  \phantom{00} 2000    &    &  \phantom{00} 64     &  & numeric  &  \phantom{000}10\\
           5\phantom{0000}&     DNA         &  \phantom{00} 3186    &    &  \phantom{00}180     &  & nominal  &  \phantom{0000}3\\
           6\phantom{0000}&     isolet5    &  \phantom{00} 1559     &    &  \phantom{00} 617     &  & mixed  & \phantom{000}26\\
           7\phantom{0000}&     Colon Tumor    &  \phantom{00} 62    &    &  \phantom{00} 2000    &  & numeric  &  \phantom{0000}2\\
           8\phantom{0000}&     BCR\_ABL   &  \phantom{00} 112    &    &  \phantom{00} 12559     &  & numeric  &  \phantom{0000}2\\
           9\phantom{0000}&     Prostate Cancer      &  \phantom{00} 34    &    &  \phantom{00} 12601    &  & numeric  &  \phantom{0000}2\\
           10\phantom{0000}&     Breast Cancer      &  \phantom{00} 19    &    &  \phantom{00} 24482    &  & numeric  &  \phantom{0000}2\\
            \bottomrule
        \end{tabular}
    \end{threeparttable}
\end{table}

\subsection{Classifiers and Experimental settings}
\subsubsection{Classifiers}
In our experiments, four famous and most frequently used classifiers -- Na\"{i}ve Bayesian Classifier (NBC) \cite{Weka}, Support Vector Machine (SVM) \cite{SVM}, $k$-Nearest Neighbor ($k$NN) \cite{kNN} and C4.5 decision tree \cite{5} are adopted to generate classification error rate on the datasets with selected features preprocessed by different feature selection methods. We set $k=1$ for $k$NN and employ Gaussian RBF kernels for SVM.

\subsubsection{Experimental settings}
First, we show the classification results of the four classifiers on $1,...,m$ selected features for each feature selection method, where $m$ in our experiments is set to be
$\min\left\{50,\left\lfloor\frac{{|\s{F}|}}{2}\right\rfloor\right\}.$
10-fold cross validation is applied in this part. Note that the nature of the learning process of each classifier is different. Since we are
interested in checking the quality of the selected
features, independently from the type of classification rule applied, the average result of the four classifiers is thus reported.

In addition, we compare the best classification results for the six feature selection methods among their selected features. That is, we check the average classification results for each feature selection method on the datasets with selected features ranging from 1 to $\min\left\{50,\left|\s{F}\right|\right\}$, and report the best one.
 In order to achieve stable results, a $(M=10)\times (N=10)$-fold cross-validation is applied, i.e. 10-fold validation will be conduct ten times for each classifier on each dataset. Thus, a total of one hundred result samples (i.e. average results from four classifiers) can be collected where each sample is an average classification result of the four classifiers. Finally, the average of one hundred samples is reported in our paper. Wilcoxon rank-sum test is applied to determine the statistical significance of the difference of the results (where the significant level is set to be 0.05).

At last, to test the stability of the performance on different datasets, average classification results of different datasets, in ranges from 1 to 5, from 1 to 10, from 1 to 15, from 1 to 20, from 1 to 25, from 1 to 30, from 1 to 35, from 1 to 40,from 1 to 45, and from 1 to 50 selected features, are reported and analyzed respectively for each classifier and feature selection method. Friedman test is applied to analyze the statistical significance of the results. These ten average classification results have been considered to be the approximate transitory period
to reach a stable performance for the datasets used.

\subsection{Experimental results and discussion}\label{ExperimentalResults}

\begin{figure*}[!ht]
\begin{center}
  \includegraphics[ scale = 0.6]{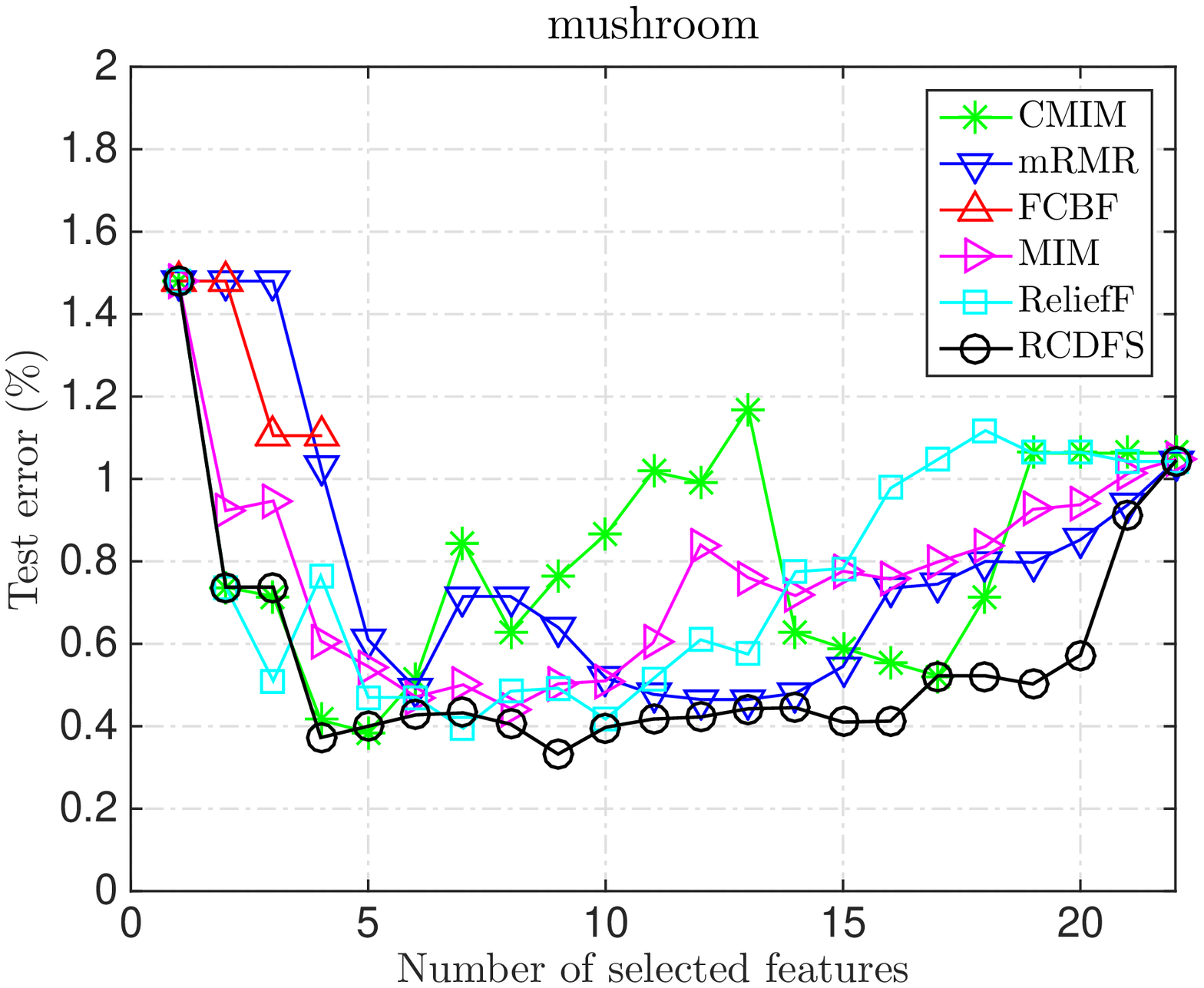}\\
  \caption{Accuracy comparison with different number of selected features on mushroom.}\label{mushroom}
\end{center}
\end{figure*}

\begin{figure*}[!ht]
\begin{center}
  \includegraphics[ scale = 0.6]{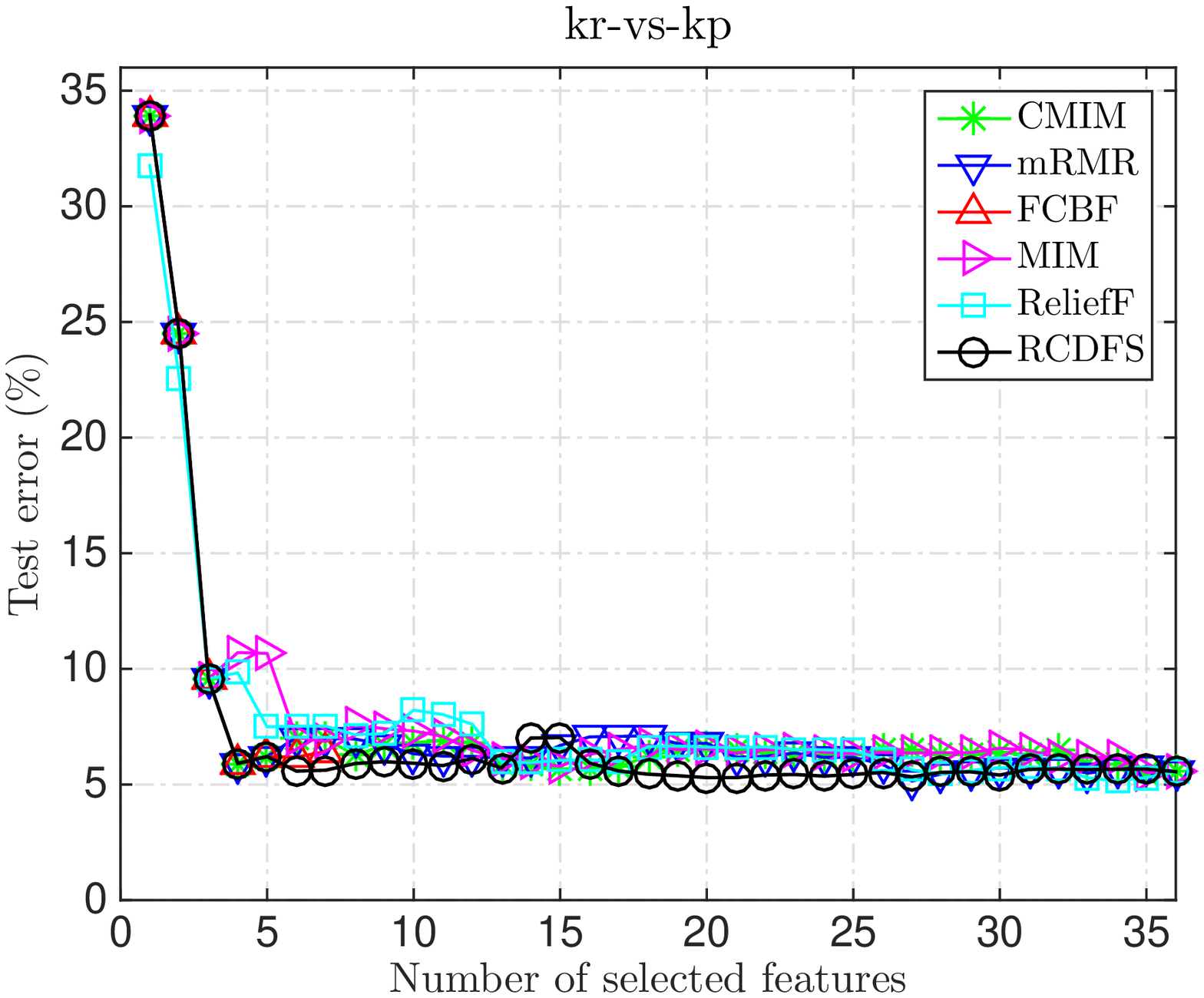}\\
  \caption{Accuracy comparison with different number of selected features on kr-vs-kp.}\label{kr-vs-kp}
\end{center}
\end{figure*}

\begin{figure*}[!ht]
\begin{center}
  \includegraphics[scale = 0.6]{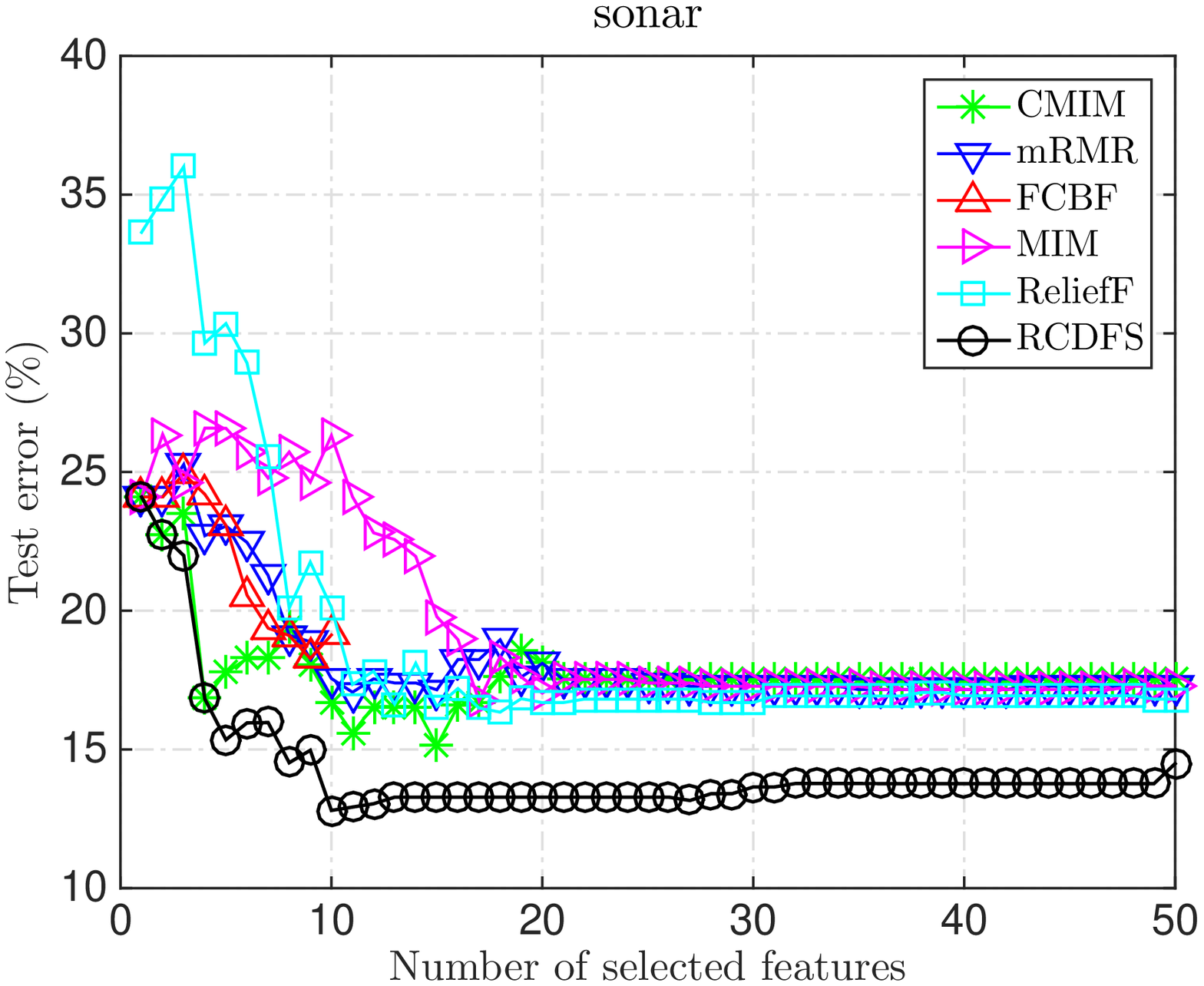}\\
  \caption{Accuracy comparison with different number of selected features on sonar.}\label{sonar}
\end{center}
\end{figure*}

\begin{figure*}[!ht]
\begin{center}
  \includegraphics[trim = 0 -5em 0 -7.5em, scale = 0.6]{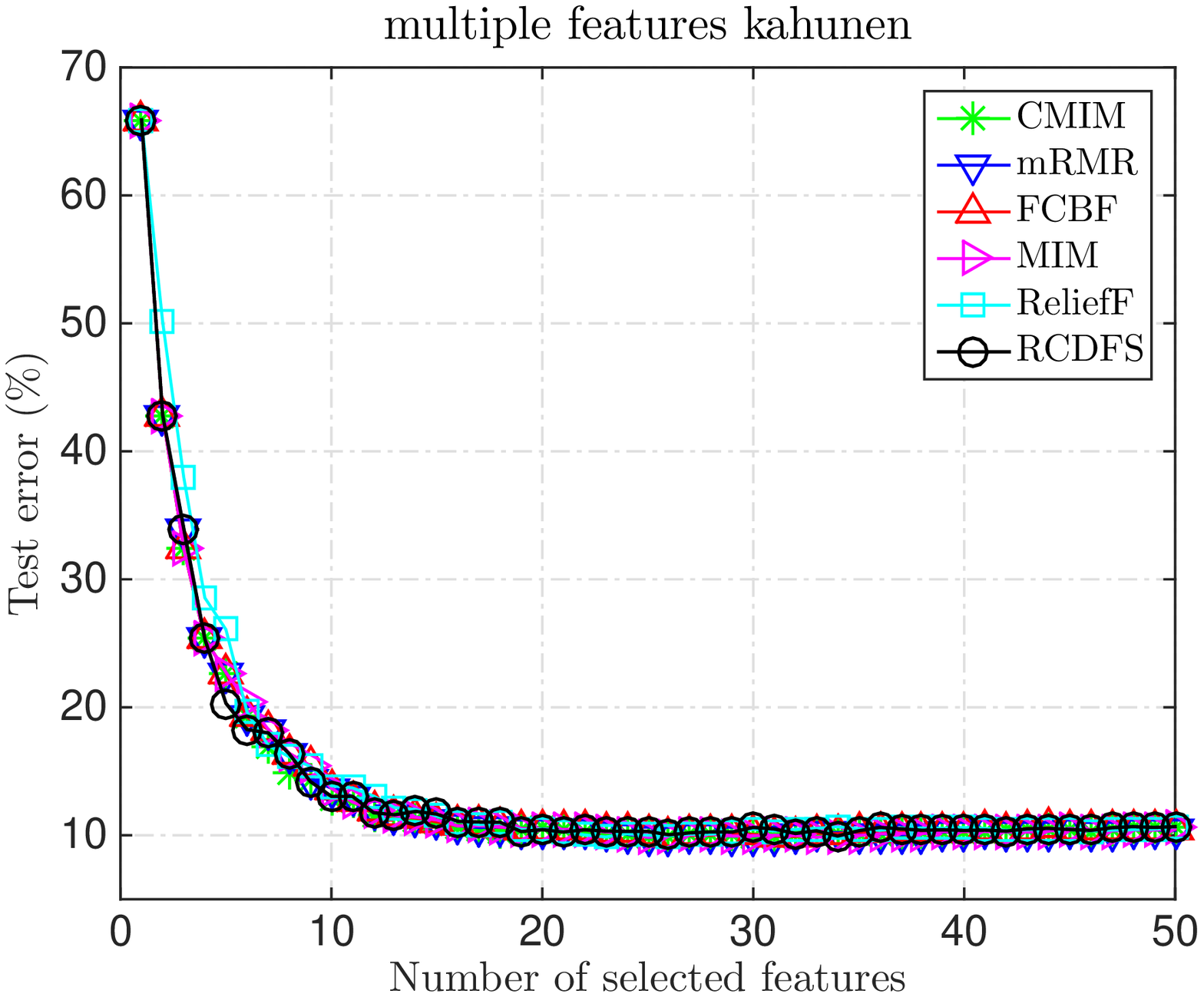}\\
  \caption{Accuracy comparison with different number of selected features on multiple features kahunen.}\label{multiple features kahunen}
\end{center}
\end{figure*}

\begin{figure*}[!ht]
\begin{center}
  \includegraphics[scale = 0.6]{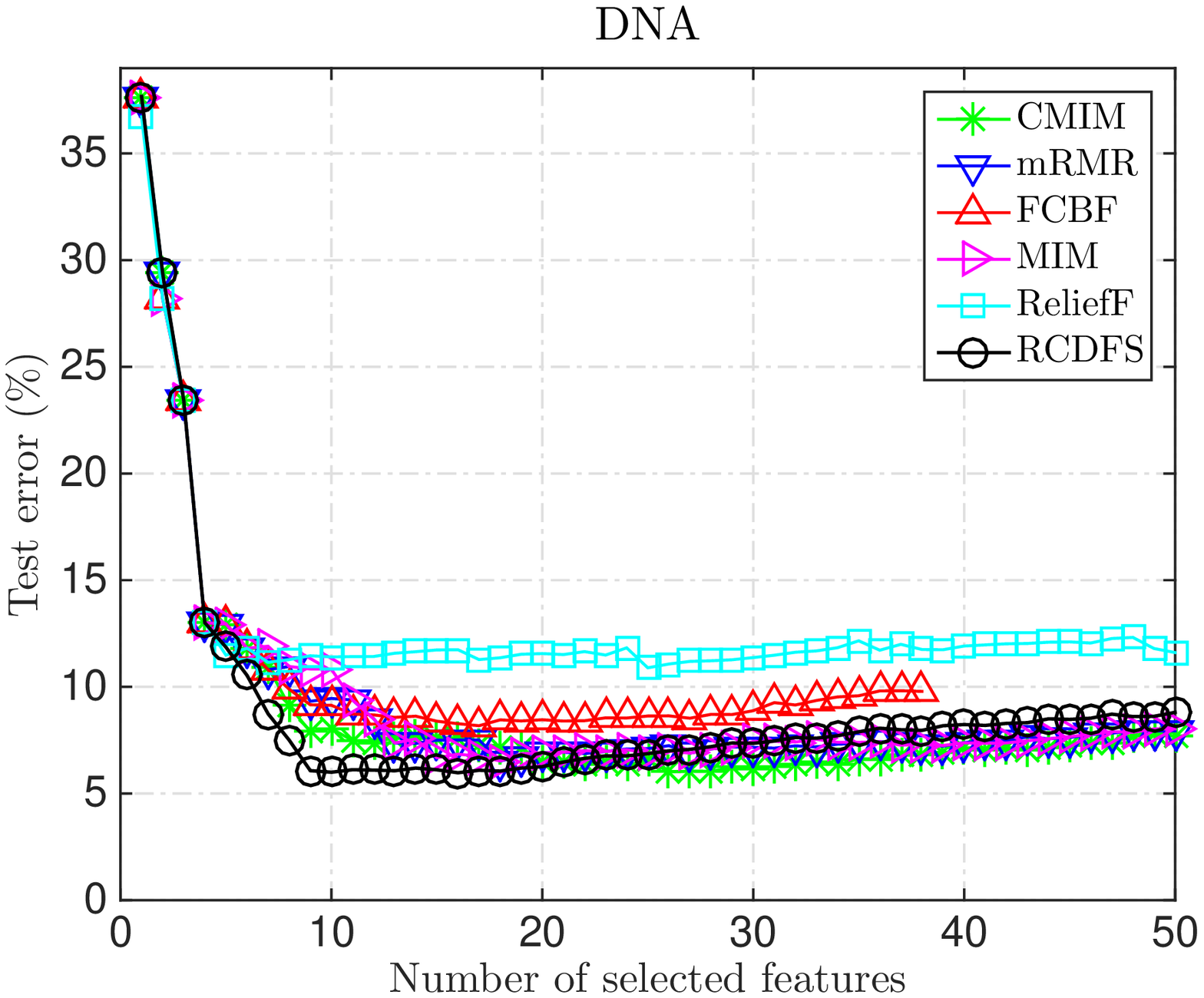}\\
  \caption{Accuracy comparison with different number of selected features on DNA.}\label{DNA}
\end{center}
\end{figure*}

\begin{figure*}[!ht]
\begin{center}
  \includegraphics[scale = 0.6]{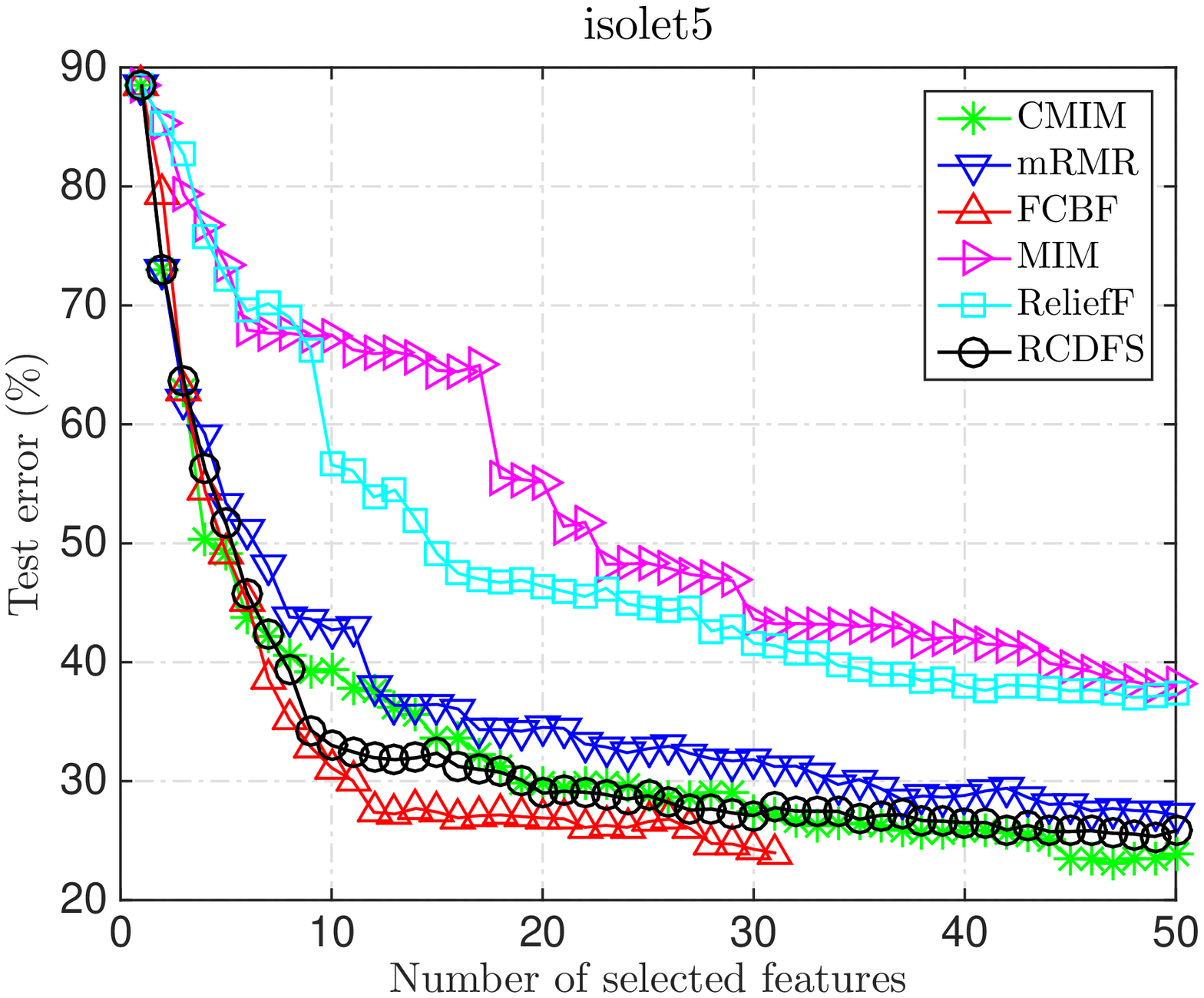}\\
  \caption{Accuracy comparison with different number of selected features on isolet5.}\label{isolet5}
\end{center}
\end{figure*}

\begin{figure*}[!ht]
\begin{center}
  \includegraphics[scale = 0.6]{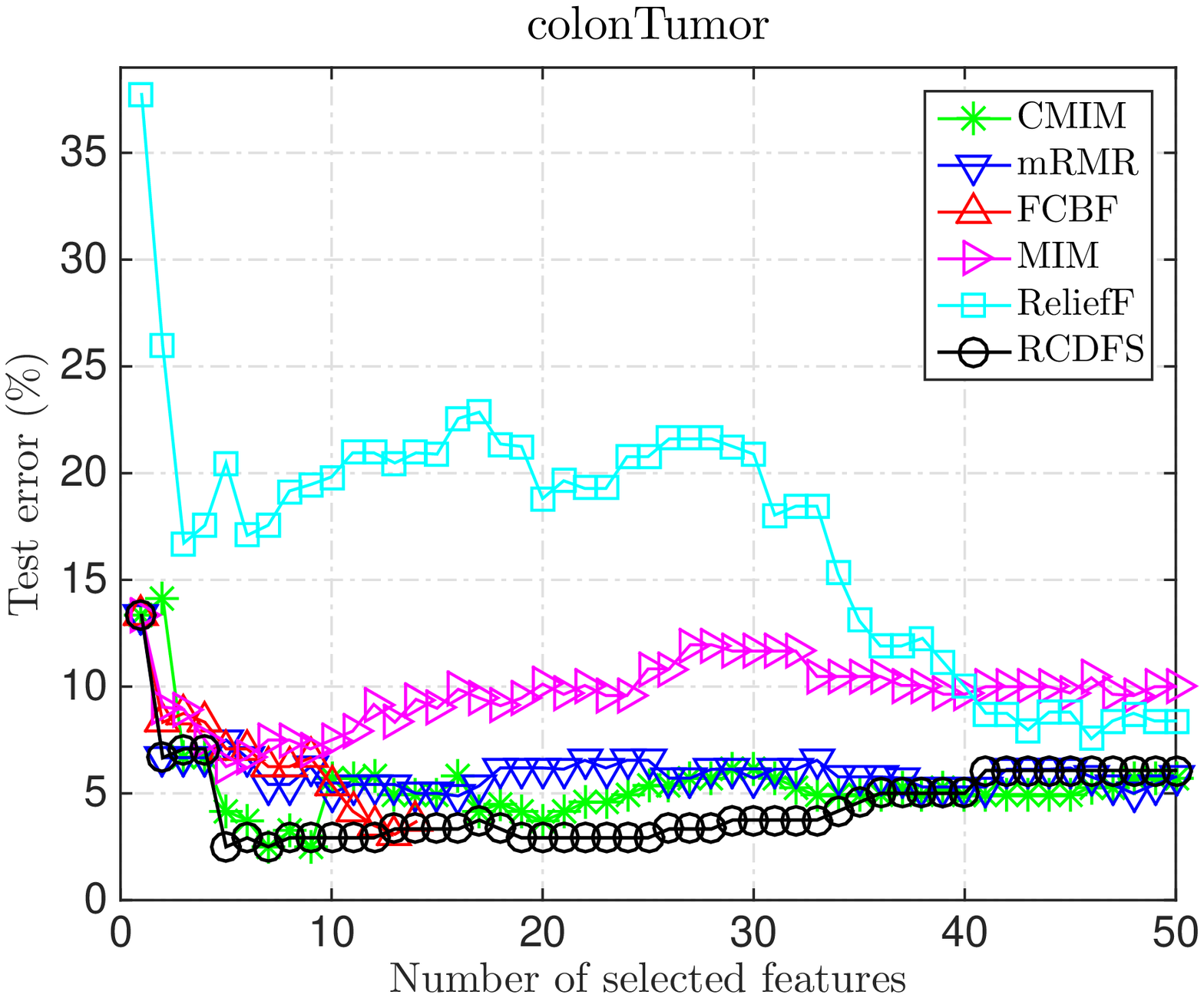}\\
  \caption{Accuracy comparison with different number of selected features on Colon Tumor.}\label{ColonTumor}
\end{center}
\end{figure*}

\begin{figure*}[!ht]
\begin{center}
  \includegraphics[scale = 0.6]{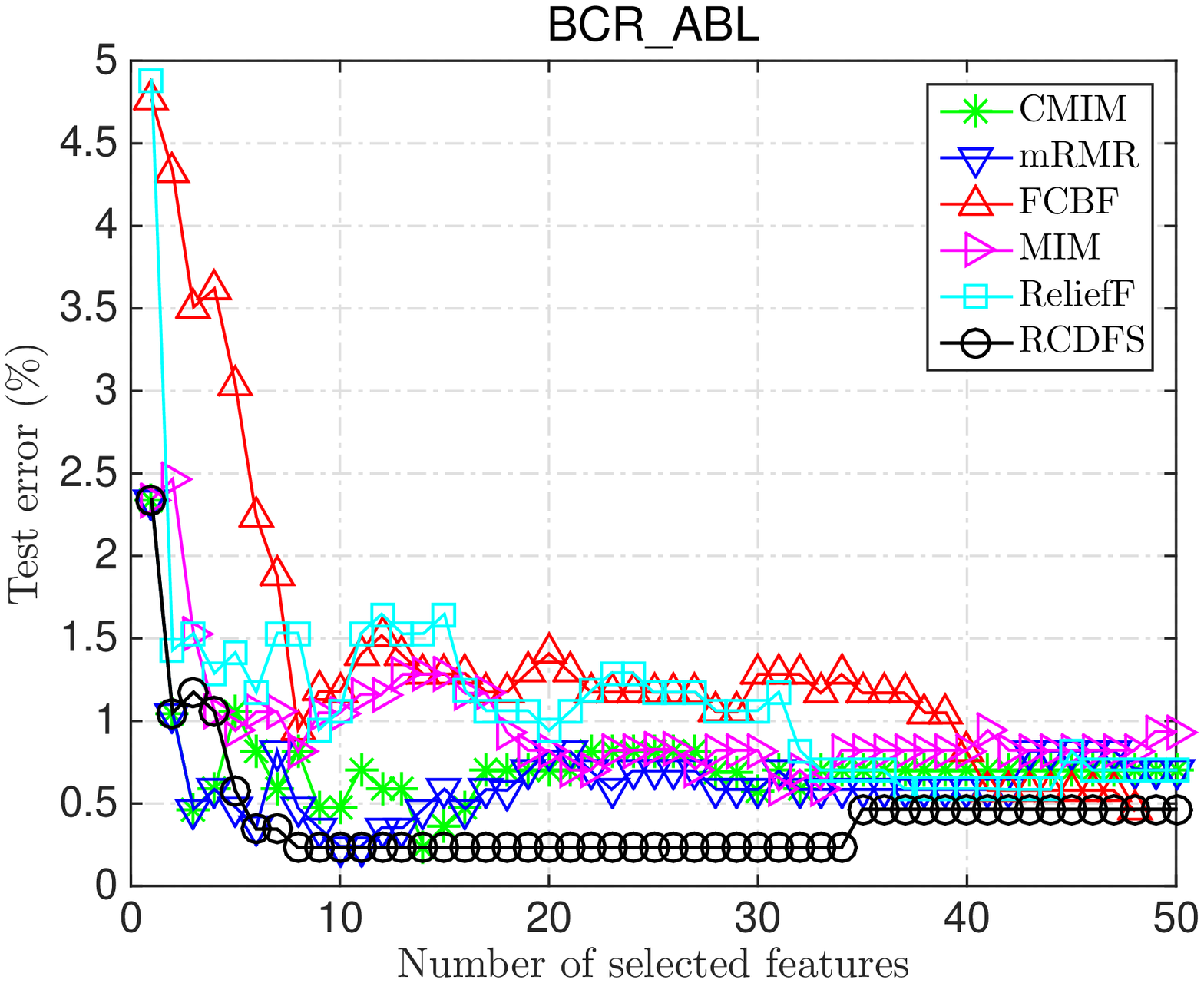}\\
  \caption{Accuracy comparison with different number of selected features on BCR\_ABL.}\label{BCR_ABL}
\end{center}
\end{figure*}

\begin{figure*}[!ht]
\begin{center}
  \includegraphics[scale = 0.6]{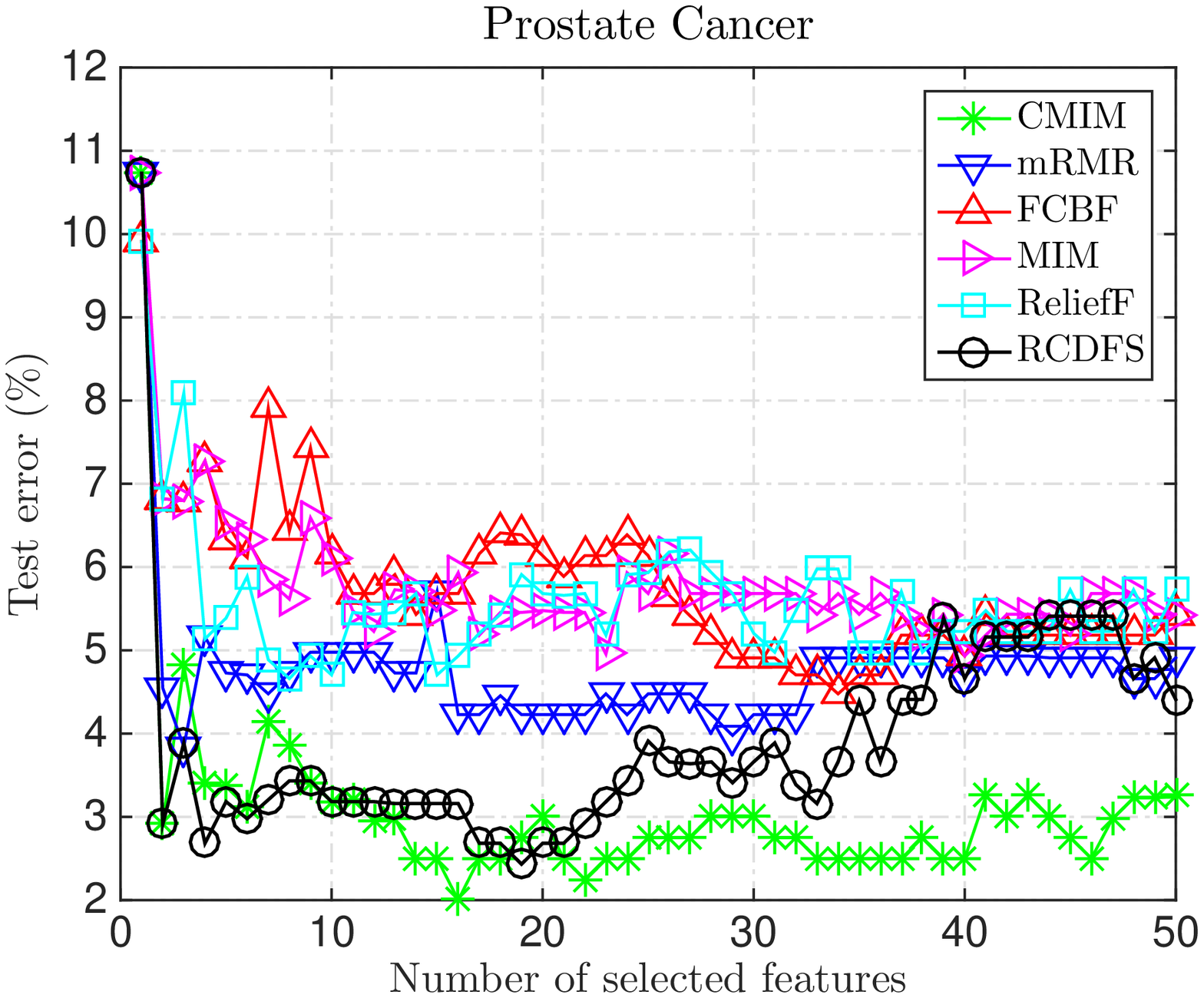}\\
  \caption{Accuracy comparison with different number of selected features on Prostate Cancer.}\label{ProstateCancer}
\end{center}
\end{figure*}

\begin{figure*}[!ht]
\begin{center}
  \includegraphics[scale = 0.6]{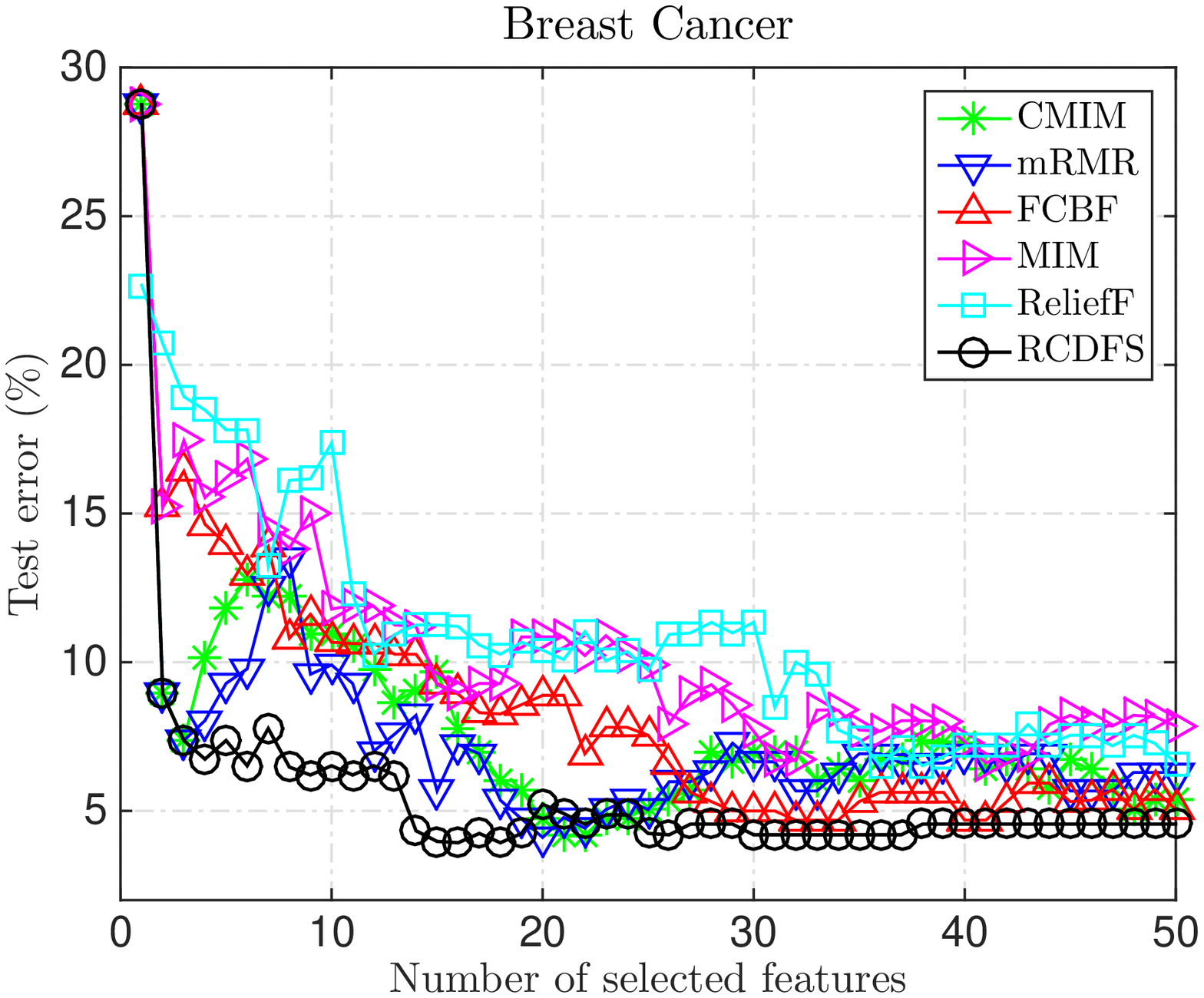}\\
  \caption{Accuracy comparison with different number of selected features on Breast Cancer.}\label{BreastCancer}
\end{center}
\end{figure*}

 Figs.\ref{mushroom}--\ref{BreastCancer} show the 10-fold cross-validation average test error rate of the different types of classifiers (NBC, SVM, kNN, and C4.5) on the ten datasets to illustrate the effectiveness of proposed method RCDFS, where the consecutive numbers of selected features are described by X axis, and the average test error rate is represented by Y axis. According to the results shown in Figs.\ref{mushroom}--\ref{BreastCancer}, the superiority of RCDFS can be verified in the majority of cases. Particularly on seven datasets namely mushroom (Fig.\ref{mushroom}), kr-vs-kp (Fig.\ref{kr-vs-kp}), sonar (Fig.\ref{sonar}), DNA (Fig.\ref{DNA}), Colon Tumor (Fig.\ref{ColonTumor}), BCR\_ABL (Fig.\ref{BCR_ABL}), and Breast Cancer (Fig.\ref{BreastCancer}), RCDFS significantly outperforms CMIM, mRMR, FCBF, MIM, and ReliefF. More precisely, RCDFS usually perform better at the beginning of feature selection process on several datasets such as sonar (Fig.\ref{sonar}) and Colon Tumor (Fig.\ref{ColonTumor}). This is probably because the redundancy and complementariness are both considered by RCDFS, rather than only measuring pairwise redundancy like mRMR and CMIM or ignoring redundancy among features like MIM, FCBF and ReliefF. For other datasets, e.g. BCR\_ABL dataset, the test error rate corresponding to RCDFS is higher than that to CMIM and mRMR on the first five selected features, whereas after the sixth feature being selected, RCDFS performs better (i.e. the test error is lower) than other methods and it is never exceeded, which is possibly due to the fact that the dispersion of redundancy-complementariness correlation becomes influential to feature evaluation process after several features being selected, i.e. the interference effect of FPs in the selected subset impairs the evaluation ability of the selected compared algorithms except for RCDFS. On the whole, it can also be seen that RCDFS selects less features corresponding to the lowest error rate than other methods (e.g. it corresponds to the best classification results only selecting five and sixteen features on Colon Tumor and DNA, respectively).
It is also found that the performance of RCDFS is not always outstanding and sometimes inferior to CMIM (such as Fig.\ref{ProstateCancer}). This may also lie in the dispersion of the redundancy-complementariness correlation since there also exist alternative conditions leading to high dispersion rather than the variance between TPs and FPs.

Tab.\ref{TabNumfeatures} records the number of features selected by each feature selection algorithm. We observe from the table that the average number of selected features of RCDFS (18.7) is smaller compared to other algorithms used in our experiment. This indicates the advantage of RCDFS that the best classification result can be obtained with a sufficiently small set of features.

\begin{table*}[!ht]
	\centering
	\footnotesize
	\begin{threeparttable}
		\caption{Number of selected features corresponding to best performance}
		\label{TabNumfeatures}
		\begin{tabular}{l rrrrrr}
			\toprule
			\multirow{2}{*}{ Datesets }
			&\multicolumn{6}{c}{\raisebox{0pt}[1mm][1mm]{\# features}}  \\
			\cline{2-7}
			& \raisebox{-2pt}[1mm][1mm]{\phantom{aa}RCDFS} & \raisebox{-2pt}[1mm][1mm]{\phantom{aa}CMIM}& \raisebox{-2pt}[1mm][1mm]{\phantom{aa}mRMR} & \raisebox{-2pt}[1mm][1mm]{\phantom{aa}FCBF} & \raisebox{-2pt}[1mm][1mm]{\phantom{aa}MIM} & \raisebox{-2pt}[1mm][1mm]{\phantom{aa}ReliefF} \\
			\midrule
			mushroom    & 9 & 5  &12 &  3  & 8  & 7 \\
			kr-vs-kp    & 21 & 35 &27 & 4 & 35 & 34  \\
			sonar       &10 & 15 & 11 & 9 & 17 & 18\\
			multiple feature kahunen  & 34 & 31 & 25 & 32 & 25 & 23\\
			DNA         & 16   &26  &18 & 17  & 18  & 25\\
			isolet5     & 49   &47  &50 & 31   &49  & 48\\
			Colon Tumor & 5 & 7 & 15 & 13 & 5 & 46 \\
			BCR\_ABL    & 8 & 14 & 10 & 48 & 31 & 42\\
			Prostate Cancer & 19  &16 &3 & 34  & 40 & 8\\
			Breast Cancer   & 16   &21 &20 & 32   & 50  & 36\\
			Avg.      & 18.7  &21.7 &19.1& 22.3 & 27.8& 28.7\\
			\bottomrule
		\end{tabular}
	\end{threeparttable}
\end{table*}

Tab.\ref{wilcoxon} show the average test error rate of NBC, SVM, $k$NN, and C$4.5$ on ten datasets over $(M=10)\times (N=10)$-fold cross validation, respectively.
For each dataset, Wilcoxon test is conducted to evaluate the statistical significance of the difference between the two groups of result samples, i.e. groups of the result samples that corresponds to RCDFS and any other feature selection method. In Tab.\ref{wilcoxon}, ``Err'' column records the average test error rate of $(M=10)\times (N=10)$-fold cross-validation. ``$p$-val'' column records the $p$-value associated with Wilcoxon test, where $p$-value less than 0.05 indicates the statistical significance of the difference between the two average values. Notation ``$\bullet$''/``$\circ$'' are used to show that the test error rate corresponding to the current feature selection method is significantly lower/higher than that to proposed method (corresponding to ``RCDFS'' column) under the test. Bold value in each row indicates that it is the best result among six feature selection methods. The average error rate of ten datasets is given in the last row.

As can be seen from Tab.\ref{wilcoxon}, the average value of test error rate for the ten datasets shows that RCDFS outperforms other methods on mushroom, sonar, DNA, BCR\_ABL, Prostate Cancer, and Breast Cancer datasets. According to the average test error rate of ten datasets given in the last row, the best one is obtained by our method ($7.89$) and the worst is by MIM ($12.57$). Also, the average test error rate of CMIM ($8.12$) is better than other algorithms (mRMR ($8.71$), FCBF ($11.1$), and ReliefF (11.62)).

\begin{figure*}[!ht]
\begin{center}
  \includegraphics[trim = 0 36em 18em 30em scale = 0.6]{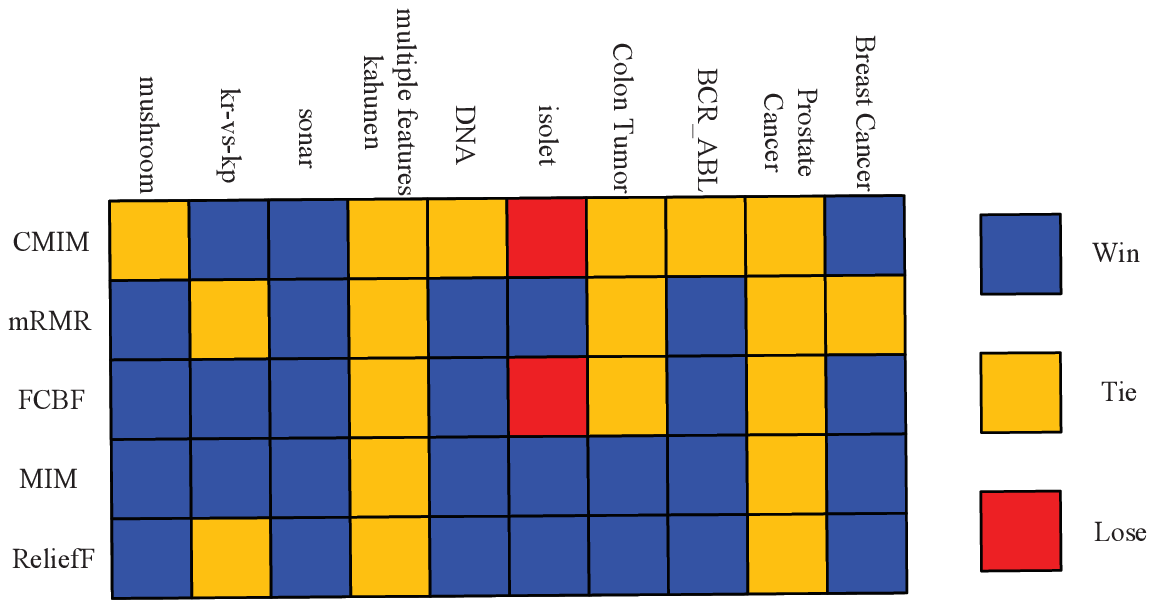}\\
  \caption{Average classification error rate comparison between RCDFS and the selected methods on the selected ten datasets.}\label{WLT}
\end{center}
\end{figure*}

For further analysis, the diagram (Fig.\ref{WLT}) is applied to visualize the statistical significance of RCDFS comparing with the selected methods under four classifiers on ten datasets. The blue box in Fig.\ref{WLT} describes that the test error rate of RCDFS is significantly better than the compared algorithms in current dataset. The yellow box represents that there is no significant difference between the results of RCDFS and the compared algorithm. The red box implies that the test error rate of RCDFS is significantly worse than the compared algorithms. The results shown in Fig.\ref{WLT} indicate that RCDFS achieves better performance in most of datasets compared with selected feature selection algorithms.

 \begin{table*}[!ht]
    \centering
 \begin{threeparttable}
 \footnotesize
   \caption{Average classification error rate of the six classifiers on selected features with NBC, SVM, kNN and C4.5, and the result of Wilocxon test.}\label{wilcoxon}
     \begin{tabular}{r c@{\hspace{3pt}} c@{\hspace{10pt}}  c@{\hspace{6pt}} r@{\hspace{1pt}} l@{\hspace{15pt}}   c@{\hspace{6pt}} r@{\hspace{1pt}} l@{\hspace{15pt}}    c@{\hspace{6pt}} r@{\hspace{1pt}} l @{\hspace{15pt}}    c@{\hspace{6pt}} r@{\hspace{1pt}} l@{\hspace{15pt}}     c@{\hspace{6pt}} r@{\hspace{1pt}} l@{\hspace{15pt}} }
     \toprule
\multirow{2}{40pt}{\# Dataset}& RCDFS&&\multicolumn{2}{c}{CMIM}&  &\multicolumn{2}{c}{mRMR}&  &\multicolumn{2}{c}{FCBF}&  &\multicolumn{2}{c}{MIM}&  &\multicolumn{2}{c}{ReliefF}&  \\
           \cline{2-2}\cline{4-5}\cline{7-8}\cline{10-11}\cline{13-14}\cline{16-17}
   &    Err        & &     Err  & $p$-val   &  &     Err  & $p$-val      &  &     Err  & $p$-val     &  &     Err  & $p$-val      &  &     Err  & $p$-val    &  \\
 \midrule
1\phantom{000} & \bf{0.32}  & & 0.37   &  0.207 &  & 0.47   &  0.000 & \fs$^\circ$  & 23.26   &  0.000 & \fs$^\circ$  & 20.57   &  0.000 & \fs$^\circ$  & 0.39   &  0.000 & \fs$^\circ$ \\

2\phantom{000} & 5.32   & & 5.61   &  0.015 & \fs$^\circ$  & \bf{5.14} &  0.231 &  & 5.91   &  0.000 & \fs$^\circ$  & 5.61   &  0.015 & \fs$^\circ$  & 5.21   &  0.449 & \\

3\phantom{000} & \bf{14.05}  & & 16.05   &  0.025 & \fs$^\circ$  & 17.44   &  0.000 & \fs$^\circ$  & 18.63   &  0.000 & \fs$^\circ$  & 17.38   &  0.001 & \fs$^\circ$  & 16.56   &  0.006 & \fs$^\circ$ \\

4\phantom{000} & 10.05   & & 9.89   &  0.317 &  & \bf{9.81} &  0.143 &  & 10.08   &  0.874 &  & \bf{9.81} &  0.143 &  & 9.83   &  0.144 & \\

5\phantom{000} & \bf{5.98}  & & 5.99   &  0.959 &  & 6.46   &  0.011 & \fs$^\circ$  & 8.16   &  0.000 & \fs$^\circ$  & 6.46   &  0.011 & \fs$^\circ$  & 10.79   &  0.000 & \fs$^\circ$ \\

6\phantom{000}& 25.13   & & \bf{23.19} &  0.000 & \fs$^\bullet$  & 27.06   &  0.000 & \fs$^\circ$  & 23.62   &  0.000 & \fs$^\bullet$  & 37.89   &  0.000 & \fs$^\circ$  & 37.11   &  0.000 & \fs$^\circ$ \\

7\phantom{000}& 3.32   & & 2.83   &  0.105 &  & 4.02   &  0.715 &  & \bf{2.37} &  0.103 &  & 7.26   &  0.001 & \fs$^\circ$  & 8.35   &  0.000 & \fs$^\circ$ \\

8\phantom{000}& \bf{5.98}  & & 5.99   &  0.959 &  & 6.46   &  0.011 & \fs$^\circ$  & 8.16   &  0.000 & \fs$^\circ$  & 6.46   &  0.011 & \fs$^\circ$  & 10.79   &  0.000 & \fs$^\circ$ \\

9\phantom{000}& \bf{5.13}  & & 5.58   &  0.070 &  & 5.71   &  0.113 &  & 5.96   &  0.960 &  & 6.71   &  0.758 &  & 8.54   &  0.053 & \\

10\phantom{000} & \bf{3.66}  & & 5.65   &  0.000 & \fs$^\circ$  & 4.52   &  0.106 &  & 4.86   &  0.021 & \fs$^\circ$  & 7.58   &  0.000 & \fs$^\circ$  & 8.63   &  0.000 & \fs$^\circ$ \\

Avg.\phantom{000} & \bf{7.89} & & 8.12 & & & 8.71 & & & 11.10 & & & 12.57 & & & 11.62 & &  \\
 \bottomrule
     \end{tabular}%
     \begin{tablenotes}
     \footnotesize
       \item $\circ$ statistical degradation at significant level of $0.05$.
       \item $\bullet$ statistical improvement at significant level of $0.05$.
     \end{tablenotes}
   \label{NBC}%
 \end{threeparttable}%
 \end{table*}

Tabs.\ref{NB_friedman}--\ref{C4.5_friedman} show the statistical significance of average error using Friedman test under different classifiers on ten datasets. Results in column $k=5$, 10, 15, 20, 25, 30, 35, 40, 45, and 50 represents the average classification error in ranges from 1 to 5, from 1 to 10, from 1 to 15, from 1 to 20, from 1 to 25, from 1 to 30, from 1 to 35, from 1 to 40,from 1 to 45, and from 1 to 50 features, respectively. Note that FCBF may select less features than other methods, e.g. it only selects 4 features on mushroom dataset, thus the average up to 4 features is described in row  $k=$ 25, 30, 35, 40, 45, and 50. A very small $p$-val (i.e. $p$-val $<0.05$) indicates the significant difference among the average values. In addition, we use S/N given in the last row of the tables to represent statistically significant/insignificant difference among the average values under Friedman test with significant level $0.05$. Bold value in each column shows the best classification result among six feature selection methods.

Tab.\ref{NB_friedman} shows that the average test error rate of Na\:{i}ve Bayesian Classifier (NB) corresponding to RCDFS is lowest among all methods and $p$-val is smaller than $0.05$. This indicates that the performance of RCDFS is best using NBC classifiers with the number of selected features in all ranges. Similar to NB, the average test error rate of SVM corresponding to RCDFS shown in Tab.\ref{SVM_friedman} is also lowest with the number of selected features in most of the ranges. In addition, the CMIM is superior to other methods with SVM in the ranges of $k = 1$ to $45$ and to $50$. From Tab.\ref{kNN_friedman} and Tab.\ref{C4.5_friedman}, the average error rates of $k$NN and C4.5 corresponding to RCDFS are both the lowest and the $p$-val is also smaller than $0.05$, which verifies the effectiveness of proposed method.

\begin{table*}[!ht]
 \centering
 \begin{threeparttable}
  \caption{Average classification error rate for all databases with NB classifiers, and the result of the Friedman test.}
  \label{NB_friedman}
 \begin{tabular}{ccccccccccc}
 \toprule
NB &$k=5$ & $k=10$ & $k=15$ & $k=20$ & $k=25$ & $k=30$ &  $k=35$ & $k=40$ & $k=45$ & $k=50$  \\
  \midrule
RCDFS &\bf{18.91}  & \bf{13.92}  & \bf{11.66}  & \bf{10.45}  & \bf{10.56}  & \bf{9.93}  & \bf{9.48}  & \bf{8.87}  & \bf{8.59}  & \bf{8.34}  \\
CMIM & 19.34  &  14.70  &  12.68  &  11.43  &  11.48  &  10.88  &  10.40  &  9.82  &  9.51  &  9.25  \\
mRMR & 19.42  &  15.32  &  13.23  &  12.07  &  12.30  &  11.73  &  11.32  &  10.88  &  10.57  &  10.29  \\
FCBF & 20.26  &  15.51  &  13.22  &  12.04  &  12.30  &  11.60  &  11.26  &  10.72  &  10.59  &  10.48  \\
MIM & 27.01  &  21.11  &  18.64  &  17.11  &  17.38  &  16.50  &  15.78  &  15.52  &  15.00  &  14.56  \\
ReliefF & 29.28  &  23.04  &  19.99  &  18.35  &  18.95  &  18.02  &  17.29  &  17.08  &  16.43  &  15.87  \\
$p$-val & 0.000  & 0.000  & 0.000  & 0.000  & 0.000  & 0.000 & 0.000 & 0.000 & 0.001 & 0.000\\
 & S & S & S & S & S & S & S & S & S & S\\
 \bottomrule
   \end{tabular}
 \end{threeparttable}
 \end{table*}

\begin{table*}[!ht]
 \centering
 \begin{threeparttable}
  \caption{Average classification error rate for all databases with SVM classifiers, and the result of the Friedman test.}
  \label{SVM_friedman}
 \begin{tabular}{ccccccccccc}
 \toprule
SVM &$k=5$ & $k=10$ & $k=15$ & $k=20$ & $k=25$ & $k=30$ &  $k=35$ & $k=40$ & $k=45$ & $k=50$  \\
  \midrule
RCDFS &\bf{18.83}  & \bf{13.65}  & \bf{11.41}  & \bf{10.15}  & \bf{10.34}  & \bf{9.72}  & \bf{9.32}  & \bf{9.34}  &  9.24  &  9.20  \\
CMIM & 18.87  &  14.24  &  12.14  &  10.75  &  10.85  &  10.12  &  9.57  &  9.37  & \bf{8.96}  & \bf{8.64}  \\
mRMR & 19.64  &  15.20  &  12.83  &  11.38  &  11.57  &  10.82  &  10.25  &  10.08  &  9.71  &  9.41  \\
FCBF & 20.81  &  15.80  &  13.49  &  12.20  &  12.46  &  11.70  &  11.29  &  10.78  &  10.59  &  10.43  \\
MIM & 27.28  &  21.15  &  18.53  &  16.60  &  16.79  &  15.61  &  14.67  &  14.75  &  14.06  &  13.49  \\
ReliefF & 29.10  &  22.17  &  18.68  &  16.52  &  16.86  &  15.82  &  14.92  &  15.09  &  14.40  &  13.87  \\
$p$-val & 0.003  & 0.000  & 0.000  & 0.000  & 0.000  & 0.001 & 0.002 & 0.001 & 0.001 & 0.002\\
 & S & S & S & S & S & S & S & S & S & S\\
 \bottomrule
   \end{tabular}
 \end{threeparttable}
 \end{table*}

\begin{table*}[!ht]
 \centering
 \begin{threeparttable}
  \caption{Average classification error rate for all databases with $k$NN classifiers, and the result of the Friedman test.}
  \label{kNN_friedman}
 \begin{tabular}{ccccccccccc}
 \toprule
$k$NN &$k=5$ & $k=10$ & $k=15$ & $k=20$ & $k=25$ & $k=30$ &  $k=35$ & $k=40$ & $k=45$ & $k=50$  \\
  \midrule
RCDFS &\bf{18.72}  & \bf{14.27}  & \bf{12.38}  & \bf{11.22}  & \bf{11.62}  & \bf{11.11}  & \bf{10.77}  & \bf{11.19}  & \bf{11.01}  & \bf{10.89}  \\
CMIM & 19.10  &  14.90  &  13.04  &  12.02  &  12.59  &  12.21  &  11.87  &  12.37  &  12.19  &  12.03  \\
mRMR & 19.56  &  15.61  &  13.68  &  12.64  &  13.22  &  12.64  &  12.21  &  12.63  &  12.38  &  12.17  \\
FCBF & 20.54  &  16.45  &  14.52  &  13.51  &  14.22  &  13.70  &  13.42  &  13.27  &  13.07  &  12.92  \\
MIM & 26.94  &  21.28  &  18.88  &  17.40  &  18.14  &  17.36  &  16.74  &  17.44  &  16.97  &  16.57  \\
ReliefF & 28.98  &  22.81  &  19.87  &  18.05  &  18.72  &  17.91  &  17.29  &  18.02  &  17.49  &  17.04  \\
$p$-val & 0.003  & 0.000  & 0.000  & 0.000  & 0.000  & 0.000 & 0.000 & 0.000 & 0.000 & 0.000\\
 & S & S & S & S & S & S & S & S & S & S\\
 \bottomrule
   \end{tabular}
 \end{threeparttable}
 \end{table*}

\begin{table*}[!ht]
 \centering
 \begin{threeparttable}
  \caption{Average classification error rate for all databases with C4.5 classifiers, and the result of the Friedman test.}
  \label{C4.5_friedman}
 \begin{tabular}{ccccccccccc}
 \toprule
C4.5 &$k=5$ & $k=10$ & $k=15$ & $k=20$ & $k=25$ & $k=30$ &  $k=35$ & $k=40$ & $k=45$ & $k=50$  \\
  \midrule
RCDFS &\bf{19.23}  & \bf{15.50}  & \bf{13.84}  & \bf{12.91}  & \bf{13.83}  & \bf{13.50}  & \bf{13.27}  & \bf{14.22}  & \bf{14.13}  & \bf{14.08}  \\
CMIM & 19.50  &  16.48  &  15.04  &  14.12  &  15.10  &  14.78  &  14.53  &  15.51  &  15.40  &  15.29  \\
mRMR & 20.26  &  17.72  &  16.19  &  15.33  &  16.36  &  15.97  &  15.68  &  16.83  &  16.71  &  16.57  \\
FCBF & 21.34  &  17.91  &  16.57  &  16.05  &  17.40  &  17.18  &  17.07  &  17.57  &  17.56  &  17.57  \\
MIM & 27.47  &  22.76  &  20.84  &  19.34  &  20.37  &  19.57  &  18.91  &  20.13  &  19.77  &  19.52  \\
ReliefF & 29.65  &  24.22  &  21.95  &  20.68  &  21.94  &  21.32  &  20.69  &  21.90  &  21.29  &  20.81  \\
$p$-val & 0.019  & 0.000  & 0.000  & 0.000  & 0.000  & 0.000 & 0.001 & 0.000 & 0.000 & 0.000\\
 & S & S & S & S & S & S & S & S & S & S\\
 \bottomrule
   \end{tabular}
 \end{threeparttable}
 \end{table*}

\section{Conclusions \& future work}\label{conclusion}
Relevance and redundancy are two important feature properties attracting much attention in the study of feature selection. Many algorithms eliminate redundancy by measuring pairwise inter-correlation between features, while they cannot identify the complementariness of features and the correlation among more than two features. Although the former problem can be effectively addressed by introducing a modification item, high inter-correlation of features still makes the result far from optimal. Specifically, pairwise approximation of high inter-correlation may misidentify and select FPs which will in turn impair the effectiveness of feature evaluation. In order to identify the interference effect of FPs, the redundancy-complementariness dispersion is taken into account in proposed method to adjust the measurement of pairwise inter-correlation of features.
To illustrate the effectiveness of proposed method RCDFS, classification experiments are conducted with four frequently used classifiers on ten datasets. In the experiments, RCDFS is compared with five representative feature selection methods namely CMIM, mRMR, FCBF, MIM, and ReliefF. Classification results have been proven to perform satisfactorily of RCDFS. To verify the stability of RCDFS, Wilcoxon test as well as Friedman test are adopted to assess the statistical significance of the differences among the results of the feature selection method. According to the test results, RCDFS performs better than the selected methods in most of the cases.

Although the superiority of RCDFS has been verified in the experiments, there still remain challenges which are imperative to be solved in our future work. One is that how to properly set the weights of three objectives, i.e. coordinate relevance, redundancy-complementary, and dispersion of pairwise inter-correlation, is still needed to be studied. Possible directions include multi-objective programming and multi-index evaluation techniques such as data envelopment analysis. Additionally, since their is no causal relationship between FPs and the dispersion of pairwise inter-correlation, only concerning such dispersion may not always be effective in feature evaluation. Thus how to design more effective heuristics in the context of first-order approximation will be further studied in future.

\section*{Acknowledgement}

The corresponding author would like to thank the support from the Doctorate Fellowship Foundation of Huazhong University of Science \& Technology (D201177780), the Fundamental Research Funds for the Central Universities, HUST (CXY12Q044, CXY13Q035), the Graduates' Innovation Fund of Huazhong University of Science \& Technology (HF-11-20-2013), and China Scholarship Council (201406160046).

\bibliographystyle{elsarticle-num}
\hspace*{\stretch{1}}
\bibliography{bibdata}







\end{document}